\documentclass[authoryear,3p,times,twocolumn]{elsarticle}
\usepackage{amsthm}
\usepackage{footnote}
\usepackage{amsmath}
\usepackage{algpseudocode}
\usepackage[ruled, vlined, linesnumbered, resetcount]{algorithm2e}
\usepackage{subfig}
\usepackage{placeins}
\usepackage[normalem]{ulem}
\usepackage{enumitem}
\usepackage{multirow}
\useunder{\uline}{\ul}{}
 \usepackage[draft]{hyperref}

\journal{X}

\begin{document}

\begin{frontmatter}



\title{On the Reliable Detection of Concept Drift from Streaming Unlabeled Data}


\author[tj]{Tegjyot Singh Sethi \corref{cor1}}
\ead{tegjyotsingh.sethi@louisville.edu}
\author[tj]{Mehmed Kantardzic}
\ead{mehmedkantardzic@louisville.edu}


\address[tj]{Data Mining Lab, University of Louisville, Louisville, USA }
\cortext[cor1]{Corresponding author.}

\begin{abstract}

Classifiers deployed in the real world operate in a dynamic environment, where the data distribution can change over time. These changes, referred to as concept drift, can cause the predictive performance of the classifier to drop over time, thereby making it obsolete. To be of any real use, these classifiers need to detect drifts and be able to adapt to them, over time. Detecting drifts has traditionally been approached as a supervised task, with labeled data constantly being used for validating the learned model. Although effective in detecting drifts, these techniques are impractical, as labeling is a difficult, costly and time consuming activity. On the other hand, unsupervised change detection techniques are unreliable, as they produce a large number of false alarms. The inefficacy of the unsupervised techniques stems from the exclusion of the characteristics of the learned classifier, from the detection process. In this paper, we propose the Margin Density Drift Detection (MD3) algorithm, which tracks the number of samples in the uncertainty region of a classifier, as a metric to detect drift. The MD3 algorithm is a distribution independent, application independent, model independent, unsupervised and incremental algorithm for reliably detecting drifts from data streams. Experimental evaluation on 6 drift induced datasets and 4 additional datasets from the cybersecurity domain demonstrates that the MD3 approach can reliably detect drifts, with significantly fewer false alarms compared to unsupervised feature based drift detectors. At the same time, it produces performance comparable to that of a fully labeled drift detector. The reduced false alarms enables the signaling of drifts only when they are most likely to affect classification performance. As such, the MD3 approach leads to a detection scheme which is credible, label efficient and general in its applicability. 

\end{abstract}

\begin{keyword}
Concept drift \sep Streaming data \sep Unlabeled \sep Margin density  \sep Ensemble\sep Cybersecurity.
\end{keyword}

\end{frontmatter}


\section{Introduction}
\label{sec:introduction}

Machine Learning has ushered in an era of data deluge, with the increasing scale and reach of modern day web applications \citep{wu2014data}. Classification has been adopted as a popular technique for providing data-driven prediction/detection capabilities, at the core of several otherwise complicated or intractable tasks. The ability to generalize and extrapolate from data has made its usage attractive as a general approach to data driven problem solving. However, the generalization ability of classifiers relies on an important assumption of \textit{Stationarity}- which states that the training and the test data should be Identically and Independently Distributed (IID), derived from the same distribution \citep{zliobaite2010change}. This assumption is often violated in the real world, where dynamic changes occur constantly. These changes in the data distribution, called Concept Drift, can cause the predictive performance of the classifiers to degrade over time. To ensure that classifiers operating in such dynamic environments are useful and not obsolete, an adaptive machine learning strategy is warranted, which can detect changes in data and then update the models as new data becomes available. 

While several adaptive techniques have been proposed in literature \citep{gama2004learning, baena2006early,bifet2007learning, goncalves2014comparative}, they rely on the unhindered and unbounded supply of human expertise, in the form of labeled data, to detect and adapt to drifting data. In a streaming environment, where data flows in constantly, such constant human intervention is impractical, as labeling is time consuming, expensive and in some cases, not a possibility at all   \citep{lughofer2016recognizing, krempl2014open}.  To highlight the problem of label dependence, consider the task of detecting hate speech from live tweets \citep{burnap2016us}, using a classification system facing the twitter stream (estimated at 500M daily tweets \footnote{\url{http://www.internetlivestats.com/twitter-statistics} (2016)}). If 0.5\% of the tweets are requested to be labeled, using crowd sourcing websites such as Amazon's Mechanical Turk\footnote{\url{www.mturk.com}}, this would imply a daily expenditure of \$50K (each worker paid \$1 for 50 tweets), and a continuous availability of ~350 crowd sourced workers (assuming each can label 10 tweets per minute, and work for 12 hours/day), every single day, for this particular task alone. The scale and velocity of modern day data applications makes such dependence on labeled data a practical and economic limitation. Streaming data applications need to be able to operate and detect drifts from unlabeled, or atmost sparsely labeled data, to be of any real use. 

\begin{figure}[t]
  \centering
  \includegraphics[width=0.95\linewidth]{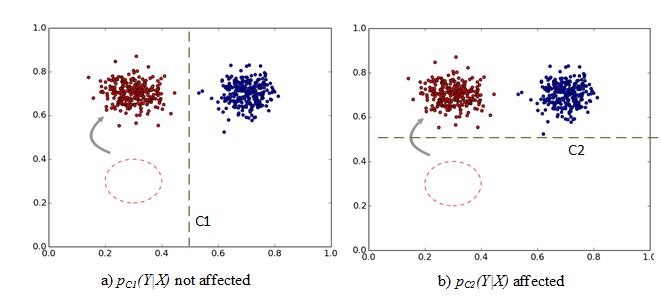}
   \caption{Drift as a function of the learned classifier model.}
  \label{fig:changes}
\end{figure}

Although the use of labeled data for retraining and updating models is largely unavoidable, its use for the purpose of drift detection is superfluous. The need for constant validation of the learned model leads to wasted labels, which are discarded when the model is found to be stable \citep{sethi2015don}. This has motivated the development of unlabeled drift detection techniques \citep{ditzler2011hellinger, da2016using}, which monitor changes to the feature distribution, as an early indicator of drift. However, existing methods using unlabeled data are essentially change detection techniques that detect any change to the data distribution, irrespective of its effects on the classification process  \citep{lee2012detection, ditzler2011hellinger, kuncheva2014pca,  qahtan2015pca, da2016using}. For the task of classification, change is relevant only when it causes model performance to degrade. This relevance is a function of the learned model, as illustrated in Figure~\ref{fig:changes}, where the same data shift resulted in diametrically opposite results. In Figure~\ref{fig:changes}a), the model performance is unaffected, while in b), there is a complete failure in the prediction capabilities of C2. The difference lies in the classier models C1 and C2, which are a result of learning on different views of the same data. The existing unlabeled techniques fail to make this distinction between the two cases, as they totally exclude the classifier from the detection process and make decisions solely on the distribution characteristics of the unlabeled data. This results in increased sensitivity to change and a large number of generated false alarms. False alarms in drift detection makes the algorithm overly paranoid and leads to wasted labeling effort, which is spent to verify relevance of the change.

\begin{figure}[t]
  \centering
  \includegraphics[width=0.95\linewidth]{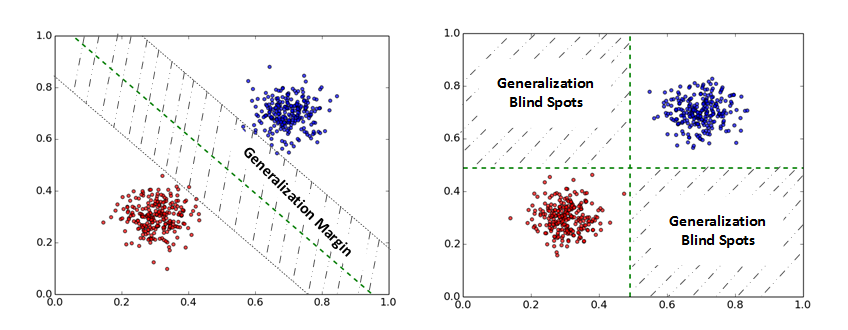}
   \caption{Classifier blindspots(margin) for SVM (left) and Feature-Bagged Ensemble (right).}
  \label{fig:blindspots}
\end{figure}

From a probabilistic perspective, concept drift can be seen as a change in the joint probability distribution of the data samples \textit{X} and their corresponding class labels \textit{Y}, as per Equation~\ref{eqn:conceptdrift} \citep{gao2007appropriate}. Unlabeled change detection techniques track changes to $P(X)$, while the labeled drift detection approaches directly track $P(Y|X)$. In this paper, an unlabeled drift detection methodology is proposed, which can vicariously track changes to $P(Y|X)$, without needing explicit labeled samples. Changes are tracked based on the distribution of sample relative to the learned classifier's boundary, to make it robust towards irrelevant changes in distribution of data. 

\begin{equation}
P(X,Y) = P(Y|X) . P(X)
\label{eqn:conceptdrift}
\end{equation}

The Margin Density Drift Detection (MD3) methodology, proposed in this paper, monitors the number of samples in a classifier's region of uncertainty (its margin), to detect drifts. Robust classifiers, such as Support Vector Machines (SVM) \citep{chang2011libsvm} or a feature bagged ensemble \citep{bry}, after training, have regions of uncertainty called margins as depicted in Figure~\ref{fig:blindspots}. These regions are a result of the classifier's attempt to generalize over unseen data and they represent the model's best guess over that data space. A large margin width with a low density (given by number of samples) , is at the core of any optimization based classification process (such as SVM). While, explicit information about class distribution is learned in the training of a classifier, an additional auxiliary information also learned and often overlooked is the margin characteristics, such as the expected margin density. This information is representative of the data state and any change in it could indicate Non-Stationarity. Margin is crucial to the generalization process and any changes to the margin density is worthy of further verification. Since the margin density can be computed from unlabeled data only, it could be used as a substitute to explicit labeled drift detection techniques, for monitoring changes in $P(Y|X)$. With this motivation, the MD3 methodology is proposed as a application independent, classifier independent, unlabeled and incremental approach to reliably signal concept drift from streaming data.

While false alarms in change detection are a hassle, due to the increased labeling expenditure and the need for frequent verification, this behavior 
is especially undesirable in cybersecurity applications because- a) Frequent false alarms annoys experts, who provide model verification, causing the detection process to loose credibility, b) An overly reactive system can be used by an adversary to manipulate learning, or to cause it to spend an excessive amount of money on labeling \citep{barreno2010security} and c) Increased labeling due to false alarms are expensive (even using crowd-sourcing websites at large scale, every day is expensive) and they cause delay in detection of attacks. As such, we will evaluate the MD3 approach as a domain independent methodology first and then also evaluate its applicability as a reliable drift detection approach in adversarial streaming domains. 

Research work in the area of dealing with concept drift \citep{sethi2016grid}, motivated the need for an unlabeled approach for drift detection, with an initial evaluation of our work presented in \citep{tsethi}. In this paper, we extend the earlier version with: i) Extending the MD3 signal as a general classifier independent approach by allowing its usage with classifiers having margins (such as SVM) and also with ensemble of classifiers (using random subspace ensembles), ii) Presenting and evaluating a novel drift induction scheme for controlled testing of effects of various drift scenarios, iii) Evaluating the MD3 approach on cybersecurity domain datasets and benchmark concept drift datasets, demonstrating generality and efficacy of the approach irrespective of the underlying classifier used, and iv) Performing additional sensitivity analysis and comparison with state of the art margin based unlabeled drift detection techniques. The main contributions of this paper, to the state of the art, are: 

\begin{itemize}
\item Extending the notion of classifier uncertainty, as a uni-variate signal, for detecting concept drift. 

\item Developing the Margin Density Drift Detection (MD3) algorithm, as an incremental streaming algorithm, for detecting drifts from unlabeled data. 

\item Formulation of margin density for classifiers with explicit margins (such as SVM) and for those without explicit margins (such as Decision Trees), with comparison and empirical evaluation demonstrating their equivalence. 

\item A novel drift induction framework for introducing concept drift into datasets, for controlled experimentation on a variety of data domains. The drift induction process provides for a reusable testing framework, by enabling the addition of concept drift to static datasets, for better experimentation and analysis of drift detection methodologies. 

\item Experimental evaluation of the MD3 framework, on datasets from cybersecurity domains, highlighting the efficacy of the  proposed approach in providing reliable and robust adversarial drift detection. 
\end{itemize}

The rest of the paper is organized as follows: Section~\ref{sec:review} presents a detailed review of existing work on drift detection, as a background to our proposed work. Section~\ref{sec:md} introduces the margin density metric as a signal for change and Section~\ref{sec:md3} develops it into a streaming algorithm. Experimental results on drift induced datasets and datasets with drift from the cybersecurity domains, is presented in Section~\ref{sec:experiment}. Additional discussion and analysis about the efficacy of the MD3 approach, with motivation for future work, is presented in Section~\ref{sec:marginanalysis}.  Conclusion and avenues for future work are presented in Section~\ref{sec:conclusion}.

\section{Review of research on concept drift detection}
\label{sec:review}

\begin{table*}[t]
\centering
\caption{Summary of drift detection approaches in literature.}
\label{tbl:driftdetectors}
\begin{tabular}{|l|l|l|}
\hline
\multirow{3}{*}{\begin{tabular}[c]{@{}l@{}}Explicit \\ drift \\  detection \\ (Supervised)\end{tabular}} & Sequential analysis & \begin{tabular}[c]{@{}l@{}} CUSUM \citep{page1954continuous},\\ PHT \citep{page1954continuous}, \\ LFR \citep{wang2015concept} \end{tabular}\\ \cline{2-3} 
 & Statistical Process Control & \begin{tabular}[c]{@{}l@{}}DDM \citep{gama2004learning},\\ EDDM  \citep{baena2006early}, \\ STEPD \citep{nishida2007detecting},\\ EWMA \citep{ross2012exponentially}\end{tabular} \\ \cline{2-3} 
 & \begin{tabular}[c]{@{}l@{}}Window based distribution \\ monitoring\end{tabular} & \begin{tabular}[c]{@{}l@{}} ADWIN \citep{bifet2007learning},\\ DoD \citep{sobhani2011new}, \\ Resampling \citep{harel2014concept} \end{tabular} \\ \hline
\multirow{3}{*}{\begin{tabular}[c]{@{}l@{}}Implicit \\ drift \\ detection\\  (Unsupervised)\end{tabular}} & \begin{tabular}[c]{@{}l@{}}Novelty detection/\\ clustering methods\end{tabular} & \begin{tabular}[c]{@{}l@{}}OLINDDA \citep{spinosa2007olindda}, \\ MINAS \citep{faria2013novelty},\\ Woo \citep{ryu2012efficient}, \\   DETECTNOD \citep{hayat2010dct}, \\ ECSMiner \citep{masud2011classification},\\ GC3 \citep{sethi2016grid}\end{tabular} \\ \cline{2-3} 
 & \begin{tabular}[c]{@{}l@{}}Multivariate distribution \\ monitoring\end{tabular} & \begin{tabular}[c]{@{}l@{}}CoC \citep{lee2012detection},\\ HDDDM \citep{ditzler2011hellinger},\\ PCA-detect \citep{kuncheva2014pca,qahtan2015pca}\end{tabular} \\ \cline{2-3} 
 &\begin{tabular}[c]{@{}l@{}} Model dependent \\ monitoring  \end{tabular}& \begin{tabular}[c]{@{}l@{}}A-distance \citep{dredze2010we},\\ CDBD \citep{lindstrom2013drift},\\  Margin \citep{dries2009adaptive}\end{tabular} \\ \hline
\end{tabular}
\end{table*}

Detecting change is essential to trigger based stream adaptation strategies. Several methods in literature have been proposed recently \cite{goncalves2014comparative}, as summarized in Table~\ref{tbl:driftdetectors}. The techniques can be divided into two categories, based on their reliance on labeled data: Explicit/Supervised drift detectors and Implicit/Unsupervised drift detectors. Explicit drift detectors rely on labeled data to compute performance metrics such as Accuracy and F-measure, which they can monitor online over time. They detect drop in performance and as such are efficient in signaling change when it matters. Implicit drift detectors rely on properties of the unlabeled data's feature values, to signal deviations. They are prone to false alarms, but their ability to function without labeling makes them useful in applications where labeling is expensive, time consuming or not available. Table~\ref{tbl:driftdetectors} shows a taxonomy of the drift detection techniques with popular techniques in each category.
 
\subsection{Explicit concept drift detection methodologies}
\label{sec:lr_explicit}

\subsubsection{Sequential analysis methodologies}
\label{sec:lr_seq}

These techniques continuously monitor the sequence of performance metrics, such as accuracy, F-measure, precision and recall; to signal a change, in the event of a significant drop in these values. The CUmulative SUM (CUSUM) approach of \citep{page1954continuous} signals an alarm when the mean of the sequence significantly deviates from 0. As per (\ref{eqn:cusum}), the CUSUM test monitors a metric\textit{ M}, at time\textit{ t}, on an incoming sample's performance $\epsilon_t$, using parameters \textit{v} for acceptable deviation and $\theta$ for the change threshold. 

\begin{equation} \label{eqn:cusum}
\begin{aligned}
M_{0}=0; \quad M_t=max{(0,M_{t-1}+\epsilon_t-v)}  
\\
if\quad { M }_{ t }>\theta \quad then \quad 'alarm' \quad and \quad { M }_{ t }=0
\end{aligned}
\end{equation}

The max function in the above equation is used to test changes in the positive direction. For a reverse effect (eg: to measure drop in accuracy), a min function can be used. This test is memory-less and can be used incrementally. A variant of this approach is the Page-Hinckley Test (PHT) \citep{page1954continuous}, which was originally developed in the signal processing domain to detect deviation from the mean of a Gaussian signal. PHT monitor the metric as an accumulated difference between its mean and current values, as shown in (\ref{eqn:pht}). 

\begin{equation} \label{eqn:pht}
\begin{aligned}
{ M }_{ 0 }=0;\quad { M }_{ t }={ M }_{ t-1 }+({ \epsilon  }_{ t }-v);\quad { M }_{ Ref }=\min { (V) } \\ if\quad { M }_{ t }-{ M }_{ Ref }>\theta \quad then\quad 'alarm'\quad and\quad { M }_{ t }=0
\end{aligned}
\end{equation}

Where, $M_0$ is the initial metric at time t = 0. $M_t$ is the current metric computed as an accumulation of the metric so far ($M_{t-1}$) and the sample's performance at time t = $\epsilon_t$. The parameter $v$ denotes acceptable deviation from mean and $\theta$ is the change detection threshold. Both the CUSUM and the PHT are best suited for univariate change detection of a sequence of performance measures, which can be tracked for online algorithms. A related statistical change detection was proposed in \citep{wang2015concept}, to deal with imbalanced streaming data, which monitors multiple performance metrics. The technique monitors the true positive rate, false positive rate, true negative rate and false positive rate, obtained from the confusion matrix of the classification. While traditional metrics of accuracy are biased towards the majority class, the confusion matrix presents a more detailed view, suitable for imbalance class problems. 

\subsubsection{Statistical Process Control based methodologies}
\label{sec:lr_spc}

The Probably Approximately Correct (PAC) learning model of machine learning states that the error rate of a trained model will decrease with increasing number of samples, if the data distribution remains stationary \citep{haussler1990probably}. The drift detection techniques based on Statistical Process Control monitor the online trace of error rates, and detects deviations based on ideas taken from control charts. A significantly increased error rate violates the PAC model and as such is assumed to be a result of concept drift. The Drift Detection Method (DDM) \citep{gama2004learning} and the Early Drift Detection Methodology (EDDM) \citep{baena2006early} are popular techniques in this category. 

The DDM approach monitors the probability of error at time \textit{t} as $p_t$ and the standard deviation as $s_t=\sqrt[]{p_t(1-p_t)/i}$. When, $p_t+s_t$ reaches its minimum value, the corresponding values are stored in $p_{min}$ and $s_{min}$. A warning is signaled when $p_t+s_t \geq p_{min}+2*s_{min}$, and a drift is signaled when $p_t+s_t \geq p_{min}+3*s_{min}$. The EDDM was developed as an extension of DDM, and was made suitable for slow moving gradual drifts, where DDM previously failed. EDDM monitors the number of samples between two classification errors, as a metric to be tracked online for drift detection. Based on the PAC model, it was assumed that, in stationary environments, the distance (in number of samples) between two subsequent errors would increase. A violation of these condition was seen to be indicative of drift. 

The Statistical Test of Equal Proportions (STEPD) \citep{nishida2007detecting} computes the accuracy of a chunk \textit{C} of recent samples and compares it with the overall accuracy from the beginning of the stream, using a chi-squares test to check for deviation. An incremental approach was proposed in  \citep{ross2012exponentially}, where the Exponentially Weighted Moving Average(EWMA) was used to signal deviation in the average error rate, in terms of the number of standard deviations from the mean. The metric \textit{M} (here, error rate) at time \textit{t} is updated as per (\ref{eqn:ewma}). 

\begin{equation} \label{eqn:ewma}
\begin{aligned}
{ M }_{ 0 }=\mu_0; \quad { M }_{ t }=\lambda *{ M }_{ t-1 }+ (1-\lambda)*\epsilon_t
\\ if\quad { M }_{ t }-\mu_0>\theta * \sigma_0 \quad then\quad 'alarm'
\end{aligned}
\end{equation}

Where, $\mu_0$ and $\sigma_0$ are mean and standard deviation obtained from the training data, by random sampling. The error rate at time \textit{t} is given by $\epsilon_t$, $\theta$ is the acceptable deviation in terms of number of standard deviation from the mean and $\lambda$ is the forgetting factor which controls the effect of previous data on the current sample. The EDDM, STEPD and EWMA, also employ the initial warning and subsequent drift signaling system as in the DDM approach. 

\subsubsection{Window based distribution monitoring methodologies}
\label{sec:lr_window}
Unlike all the methods mentioned thus far, which operate in an incremental fashion one sample at a time, window based approaches use a chunk based or sliding window approach over the recent samples, to detect changes. Deviations are computed by comparing the current chunk's distribution to a reference distribution, obtained at the start of the stream, from the training dataset \citep{bifet2007learning}. Window based approaches provide precise localization of change point, and are robust to noise and transient changes. However, they do need extra memory to store the two distributions over time. 

The Adaptive Windowing (ADWIN) algorithm of \citep{bifet2007learning} uses a variable length sliding window, whose length is computed online according to the observed changes. In case change is present, the window is shrunk and vice-versa. Whenever two large enough sub windows of the current chunk exhibit distinct averages of the performance metric, a drift is detected. Hoeffding bounds \citep{schmidt1995chernoff} are used to determine optimal change threshold and window parameters. The ADWIN methodology was shown to provide rigorous performance guarantees, efficient memory and time complexities, and freedom from having to specify cumbersome parameter values. Another window based approach, the Degree of Drift (DoD), detects drifts by computing a distance map of all samples in the current chunk and their nearest neighbors from the previous chunk \citep{sobhani2011new}. The Degree of Drift metric is computed based on a distance map and if the distance increases more than a parameter $\theta$, a drift is signaled. The Paired Learners approach of  \citep{bach2008paired} uses a pair of \textit{reactive learner}, trained on recent chunk of data, and a\textit{ stable learner}, trained on all previously seen data. Differences in accuracies between the two approaches is indicative of a drift. This disagreement (binary value) is captured in a binary valued circular list. An increase in the number of ones beyond a change threshold $\theta$ is signaled as concept drift. Drift is managed by replacing the stable model with the reactive one and setting the circular disagreement list to all zeros. 

A recently proposed permutation based method \citep{harel2014concept} relies on the observation that randomly choosing training and testing samples from a chunk of data should lead to similar accuracy of prediction, unless the window has non-stationary data. This method is based on the idea commonly used in classifier's cross validation evaluation \citep{kohavi1995study}. In cross validation, the entire training dataset is split into K bands (sequential sets of samples). In each iteration or fold of the cross validation approach, one band is chosen as the test data and the other (K-1) bands form the training dataset. Generating a model on the training dataset and then testing on the test dataset, gives the performance on that fold of the cross validation process. The same process is repeated K times and the average performance is reported. Cross validation provides a good estimate of the generalization error, when the data is stationary. The permutation approach of \citep{harel2014concept} splits the current window into two parts, to train a model on the first half and test on the second half. The window is then shuffled and the process is repeated. A significant change in the performance between the two indicates a drift, as concept drift is sensitive to the sequence of samples. This method was shown to have better precision-recall values and robustness, when compared with the DDM, the EDDM  and the STEPD methods, described in Section~\ref{sec:lr_spc}.

\subsection{Implicit drift detection methodologies}
\label{sec:implicit}

\subsubsection{Novelty detection / Clustering based methods}
\label{sec:lr_novelty}

Novelty detection methods relies on using distance and/or density information to detect previously unseen data distribution patterns. These methods are capable of identifying uncertain suspicious samples, which need further evaluation. They define an additional \textit{'Unknown'} class label to indicate that these suspicious samples do not fit the existing view of the data \citep{spinosa2007olindda}. Clustering and outlier based approaches are popular implementation strategies for detecting novel patterns, as they summarize current data and can use dissimilarity metrics to identify new samples \citep{ryu2012efficient}.

\begin{figure}[t]
  \centering
  \includegraphics[width=0.5\linewidth]{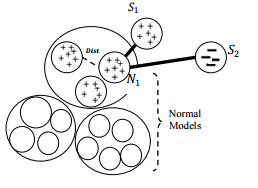}
   \caption{Distinction between Novelty (S2) and Drift (S1) in DETECTNOD \citep{hayat2010dct}.}
  \label{fig:detectnod}
\end{figure}

The OnLIne Novelty and Drift Detection Algorithm (OLINDDA), uses K-means data clustering to continuously monitor and adapt to emerging data distributions \citep{spinosa2007olindda}. Unknown samples are stored in a short term memory queue, and are periodically clustered and then either merged with existing similar cluster profiles or added as a novel profile to the pool of clusters. The MINAS algorithm of \citep{faria2013novelty}, uses micro clusters which it obtains using an incremental stream clustering algorithm- CluStream, and it extends the OLINLINDA’s approach to be used in a multi class problem. The DETECTNOD algorithm of  \citep{hayat2010dct}, uses a clustering model to define the boundaries of existing known data. It relies on Discrete Cosine Transform (DCT) to build a compact representation of these clusters, and it uses this information to provide a nearest neighbor approximation on incoming test samples. Samples falling out of the normal mode, are clustered into \textit{k} clusters and based on their similarity values to existing clusters, they are either termed as \textit{'Novelties'} or \textit{'Drifts'}. Figure~\ref{fig:detectnod} illustrates this process, where S1 is identified as a drift in the existing normal mode sub clusters, and S2 is identified as a novel pattern.

The Woo ensemble \citep{ryu2012efficient} and the ECSMiner \citep{masud2011classification} are two techniques which rely on the concept of micro clusters. The Woo ensemble clusters data and assigns a classifier to each of these clusters. A new sample falling outside the boundary of any existing cluster is marked suspicious and its density is monitored. An increased number of samples within the radius of suspicious samples indicates a new concept, which then triggers retraining of models and readjustment of the cluster centroid. The ECSMiner uses the concept of Filtered Outliers, which refers to samples which fall outside the boundary of all existing clusters, and works similar to the Woo ensemble. The GC3 approach of \citep{sethi2016grid}, extends this idea of micro clusters to be used with a grid density based clustering algorithm, where novelty is determined by newly appearing dense grids in the data space. 

All the above novelty detection techniques rely on clustering to recognize new regions of space, which were previously unseen. As such, they suffer from the curse of dimensionality, being distance dependent, and also the problem of dealing with binary data spaces. Additionally, they are suitable to detect only specific type of \textit{cluster-able} drifts. If the drift does not manifest itself as a new cluster or a novel region of space, these detection techniques will fail. Nevertheless, these techniques are suitable for multi-class classification problems, where many classes can appear and disappear during the course of the stream. 

\subsubsection{Multivariate distribution monitoring methods}
\label{sec:lr_multivariate}

Multivariate distribution monitoring approaches directly monitor the per feature distribution of the unlabeled data. These approaches are primarily chunk based and store summarized information of the training data chunk (as histograms of binned values), as the reference distribution, to monitor changes in the current data chunk. Hellinger distance and KL-divergence are commonly used to measure differences between the two chunk distributions \citep{cha2007comprehensive} and to signal drift in the event of a significant change. 

The Change of Concept(CoC) technique \citep{lee2012detection} considers each feature as an independent stream of data and monitors correlation between the current chunk and the reference training chunk. Change in the average correlation over the features is used as a signal of change. Pearson correlation was used which makes the normality assumption for the distribution. A non parametric and widely applicable unlabeled approach was proposed in \citep{ditzler2011hellinger}, called the Hellinger Distance Drift Detection Methodology (HDDDM). It is a chunk based approach which uses Hellinger distance to measure change in distribution, over time. An increased Hellinger distance, between the current stream chunk and a training reference chunk, is used to signal drift. Chunk distribution is computed by making a histogram for each feature, with $\sqrt[]{N}$ bins where \textit{N} is the number of samples in the chunk. The Hellinger distance (HD) between the reference chunk \textit{P} and the current chunk \textit{Q} is computed using Equation~\ref{eqn:hd}. Here, \textit{d} is the data dimensionality and \textit{b} is the number of bins (\textit{b}=$\sqrt[]{N}$), per feature. 

\begin{equation} \label{eqn:hd}
HD(P,Q)=  \frac { 1 }{ d } \sum _{ k=1 }^{ d }{ \sqrt { \sum _{ i=1 }^{ b }{ { \left( \sqrt { \frac { { P }_{ i,k } }{ \sum _{ j=1 }^{ b }{ { P }_{ j,k } }  }  } -\sqrt { \frac { { Q }_{ i,k } }{ \sum _{ j=1 }^{ b }{ { Q }_{ j,k } }  }  }  \right)  }^{ 2 } }  }  } 
\end{equation}

The computed Hellinger distance, which is averaged over all the features, results in a number in the range [0,$\sqrt[]{2}$]. A HD value of 0 indicates completely overlapping distributions, while $\sqrt[]{2}$ indicates total divergence. The HDDDM approach in \citep{ditzler2011hellinger} was used to detect drifts in conjunction with an incremental learning algorithm, to trigger resetting of the model. Efficacy of this approach was indicated by increased accuracy, due to interventions leading to retraining upon drift detection.  

To make the drift detection computationally efficient in high dimensional data streams, Principal Component Analysis (PCA) based feature reduction was used in \citep{kuncheva2014pca} and \citep{qahtan2015pca}. These techniques reduce the set of features to be monitored. It was shown that monitoring the reduced feature space allowed to detect drifts in the original features. \cite{kuncheva2014pca} advocates the use of the Semi Parametric Log Likelihood (SPLL) criterion to monitor changes in the data projected on the principal components. It was proposed that monitoring the principal components with the lowest 10\% of Eigenvalues is sufficient for detecting effective drifts. However, the work in \citep{qahtan2015pca} presented contrasting results. It was shown that analyzing principal components with large Eigenvalues is more valuable, as data characteristics in the original feature space are best summarized by the top component vectors, which retain the maximum variance after the reduction. 

Although the PCA based approached are efficient in reducing the number of features to be tracked, they still suffer from significant false alarms, as do the other multivariate distribution approaches. All of these methods are sensitive to changes in any of the features, irrespective of their importance to the classification task. These methods are also not suitable for detecting concept drifts in cases where drift is not manifested by feature distribution changes ($P(Y|X)$ changes but not $P(X)$, Equation~\ref{eqn:conceptdrift}). Furthermore, in the classification of imbalanced datasets these methods are not effective in tracking the changes to the minority class samples. Changes in these samples do not signal a significant deviation, as minority class samples comprise only a small percentage of the original dataset. 

\subsubsection{Model dependent drift detection methodologies}
\label{sec:lr_model}

The methodologies of Section~\ref{sec:lr_novelty} and \ref{sec:lr_multivariate} explicitly track deviations in the feature distribution of the unlabeled data. As such, they are essentially change detection methodologies which assume that a change in data distribution $P(X)$ will lead to changes in the classification performance $P(Y|X)$. While these methods are attractive for their independence to the type of classifier used, these methods lead to detecting a large number of false alarms, i.e. changes which do not lead to degradation of classification performance. False alarms lead to wasted human intervention effort and as such are undesirable. The model dependent approaches of  \citep{dries2009adaptive,lindstrom2013drift,dredze2010we,zliobaite2010change}, directly consider the classification process by tracking the posterior probability estimates of classifiers, to detect drift. They can be used with probabilistic classifiers, which output the class probabilities $P(Y|X)$ before thresholding them to generate the final class label. By monitoring the posterior probability estimates, the drift detection task is reduced to that of monitoring a univariate stream of values, making the process computationally efficient. 

The use of the Kolmogorov-Smirnov test, Wilcoxon rank sum test and the two sample t-test, was suggested in  \citep{dries2009adaptive}, to monitor the stream of posterior probability estimates. The idea of margin was introduced (using a 1-norm SVM) and the average uncertainty of samples was monitored in lieu of the multivariate feature values. This idea was extended by \citep{dredze2010we}, to the task of detecting domain shifts in high dimensional text classification applications. A reduced false positive rate was obtained by tracking the 'A-distance', which was proposed as a measure of histogram difference obtained by binning the margin distribution of samples, between the reference and current margin samples. The Confidence Distribution Batch Detection (CDBD) approach \citep{lindstrom2013drift} used KL-divergence to perform a similar analysis of classifier confidence output values (margin). Additionally, by combining it with active learning they reduce the amount of labeled data requested, over text data streams. 

These methods are attractive as they significantly reduce false alarms. However, their dependence on using probabilistic models limit their applicability. Also, these methods trigger to any change in the posterior distribution of the margin samples. Changes away from the margin of the classifier are less critical to the classification process, but none of the above mentioned approaches provide robustness against them. 

\subsection{Unlabeled drift detection in adversarial classification}
\label{sec:adversarial}

In the domain of adversarial classification, where concept drift is initiated by an attacker intended to subvert the system, unlabeled drift detection can be extremely helpful as an automated early warning system. Ensemble based techniques have been proposed in \citep{chinavle2009ensembles} and  \citep{kuncheva2008classifier}, which use disagreement scores between the ensemble models to signal changes to the data distribution. In \citep{chinavle2009ensembles}, an ensemble of classifiers is used to perform email spam classification. The average pairwise mutual agreement of the classifiers in the ensemble was used to signal change and drive retraining. However, the methodology also relies on periodic checking, using labeled samples, to ensure that high agreement is not a result of undetected changes affecting most classifiers. Recent work in \citep{smutz2016tree} also indicates the relationship between drift and classifier agreement scores. Feature bagging was found to be effective in characterizing adversarial activity in the task of malicious pdf classification. Drifts caused the classifier agreements to shift disproportionately towards the center of the [0,1] range, instead of being concentrated at the peripheries. The work in \citep{smutz2016tree} concentrates on an empirical analysis of this effect and provides initial experimentation, specific to the pdf malware domain. A  visual inspection of the change is presented. However, harnessing it as a signal for change in the context of streaming data was not explored. 

The proposed Margin Density Drift Detection (MD3) technique, in this paper, provides a way to signal drift from unlabeled data, in a reliable manner. Unlike other implicit drift detection techniques, the proposed MD3 approach is less susceptible to raising false alarms. By actively including the learned classifier's information into the decision process, the MD3 approach is able to discern changes that could adversely affect the classification performance. As such, it embodies the benefits of both the classes of drift detectors - like explicit drift detectors, it detects drifts only when they can impact classification results, and it does so using unlabeled data, saving labeling budget as with the implicit drift detectors.  In doing so, the approach bridges the gap between the two categories of drift detectors, by providing the first of its kind - domain independent, cost-effective and model independent, drift detection scheme for reliably signaling change in high dimensional data streams. 

\section{The Margin Density (MD) measure}
\label{sec:md}

\subsection{Motivation}
\label{sec:motivation}

\begin{figure}
\centering
\subfloat[Initial distribution of samples wrt. distance from classifier boundary]{\includegraphics[width=0.28\linewidth]{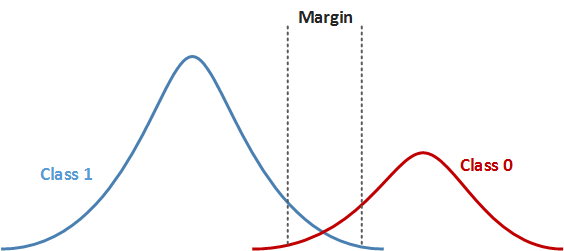}}
\hspace{1em}
\subfloat[Drift causing increase in margin density]{\includegraphics[width=0.25\linewidth]{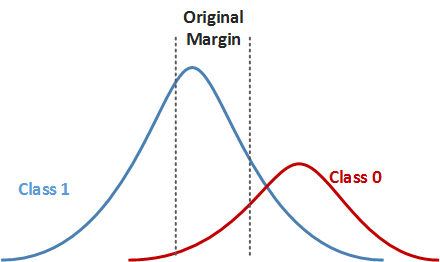}}
\hspace{1em}
\subfloat[Sudden drift causing drop in margin density ]{\includegraphics[width=0.28\linewidth]{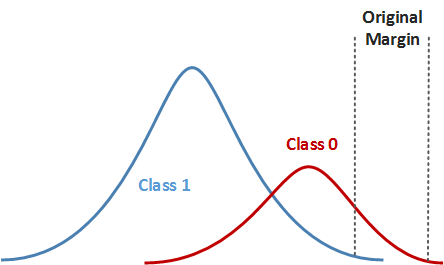}}
\caption{Drifting scenarios and their effects on the margin density.}
\label{fig:drifts}
\end{figure}

The ability to generalize from the training dataset is at the core of any classification technique. This generalization effort leads to regions of space, known as margins (Figure~\ref{fig:blindspots}), where the classifier is uncertain and tries to present a best guess based on its learned information. Margin is the portion of the prediction space which is most vulnerable to misclassification. This intuition has been used by existing works on active learning \citep{settles2010active,zliobaite2014active}, to develop labeling strategies based on uncertainty of data samples. The Uncertainty sampling technique  \citep{lindstrom2010handling} and the Query By Committee technique \citep{settles2010active} are two methodologies which select samples based on the distance from the classification boundary and the disagreement between ensemble models for a given samples, respectively. These approaches explore the informativeness of the margin samples for the task of managing labeling budget efficiently. The proposed margin density approach explores the use of margin tracking for unlabeled drift detection. 

A change in the number of samples in the margin is indicative of a drift, as depicted in Figure~\ref{fig:drifts}, where the distribution of samples with respect to their distances from the classifier boundary is shown, against a fixed margin. A sudden increase or decrease in the number of samples within the margin makes the stationary assumption of data, suspect. Classifiers define margin width and acceptable mis-classifications during the training process, to avoid over fitting. As such, tracking a sudden change in the margin characteristics can indicate distribution changes. Changes of data distribution, relative to the classification boundary, enable tracking the posterior probability distribution of the space ($P(Y|X)$) without using labeled data. This is done by tacitly involving the classifier in the detection process, thereby making the change detection process relevant to the task at hand. By using a fixed margin and by tracking the density of the samples closest to the classification boundary, false alarms caused by changes away from the boundary are avoided, as they seldom result in any performance degradation. 

The idea of margin is intuitive in probabilistic classifiers, such as logistic regression or linear Support Vector Machines, which have an explicit notion of uncertainty. However, the motivation behind its usage is more general. A classifier's boundary is an embodiment of the set of features which it deems important to the task at hand. Monitoring changes close to the boundary enables us to limit tracking to the important features only. The unlabeled drift detection approaches of Section~\ref{sec:implicit} suffer from false alarms, because they do not differentiate between changes in any of the features, giving equal weights to all features. The margin density approach tracks margin changes, which summarizes the important features and their interaction resulting in the formation of the classifier's separating boundary. 

\begin{figure}[t]
  \centering
  \includegraphics[width=0.5\linewidth]{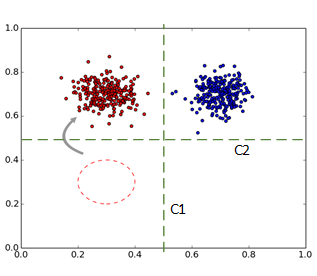}
   \caption{Change detected by coupled classifiers $C1\vee C2$. Drifted distribution leads to disagreement between C1 and C2. }
  \label{fig:coupled}
\end{figure}

In real world high dimensional datasets, there are multiple sets of features which can provide high classification performance. A robust classifier, such as a SVM with hinge loss or a feature bagged ensemble, can utilize a majority of these features, by evenly distributing weights among them, to create a better generalization over the data \citep{skurichina2002bagging,wang2015robust}. In doing so, the classifier model serves as a committee of experts with multiple independent perspectives on the same data. A change in any one perspective (set of features) will cause an increased disagreement and consequent uncertainty in the predicted results. Since, only relevant features can provide a good perspective on data, this uncertainty is indicative of a drift which requires attention. The robust classifier functions as a self-contained, self-monitoring prediction unit, aware of its own capabilities and deviations. In Figure~\ref{fig:coupled}, the coupled classifier $C1 \vee C2$ leads to a monitoring scheme where C1 and C2 are constantly monitoring each other. This coupled detection strategy can be extended to high dimensional spaces and it provides an unsupervised approach where changes in some relevant features are detected by observing invariance in other relevant features. Using this idea, the margin density approach can be extended to classifiers such as decision trees and  K- nearest neighbors, which provide explicit class labels and not probabilistic values, by using them in a feature bagged ensemble. This ensemble setup trains multiple base models, on different subset of features, and combines their results. This has the effect of distributing classification importance weights to the different features, as the features are averaged to produce the final prediction. The central idea behind the efficacy of the margin density approach lies in the ability to track changes which affect a few of the important features only. In doing so, the margin density computed from SVMs and those computed from ensemble disagreements would perform similarly, as they both track the lack of consensus and the cost of generalization, paid by the learned model.

\subsection{Computing the Margin Density (MD) metric}
\label{sec:computing}

The Margin Density metric (MD) is a univariate measure, which can be tracked over time to detect drifts from unlabeled data. Margin density is defined as:

\newtheorem{definition}{Definition}[]
\theoremstyle{definition}
\begin{definition}{\textbf{Margin Density (MD):}}
 The expected number of data samples that fall within a robust classifier's (one that distributes importance weights among its features)  region of uncertainty, i.e. its margin.
\end{definition}

The MD metric, being a ratio, has its value in the range of [0,1].  MD can be computed from classifiers with explicit notions of margins, such as a linear kernel SVM. It can also be computed by disagreement scores of feature bagged ensembles. Here, the term margin is taken as a notation for the regions of uncertainty of a robust classifier, where the classification importance weights are distributed among its features. For SVM, the margin is well defined by the algorithm, but for other classifiers, such as decision trees, we use the notion of a pseudo-margin given by the region of space with high disagreement based on a feature bagged ensemble. Section~\ref{sec:computing_svm} presents the former, while Section~\ref{sec:computing_rs}, presents the latter scenario. The term margin will be used to refer to the region of uncertainty for both the cases, as a notation.

\subsubsection{Classifiers with explicit margins}
\label{sec:computing_svm}

Classifier such as Support Vector Machines (SVM) and Logistic regression, explicitly define margins in their setup \citep{chang2011libsvm}. A soft-margin linear-kernel SVM finds an optimal maximum width separating hyperplane between two classes, by allowing a few samples to enter the margin, for better generalization capability. This is made possible by the addition of slack variables $\xi$ to the SVM's objective function, to allow for non separable cases and to add robustness against noisy samples. The optimization function of a linear kernel soft margin SVM is given by Equation~\ref{eqn:svm} \citep{chang2011libsvm}, where $w$ is the normal vector of the separating hyperplane given by $w.x+b=0$, and $b$ gives the offset from the origin, $y_i$ is the class label of the sample $x_i$ and $C$ is the regularization cost parameter which controls the misclassification cost. 

\begin{equation}
min\quad \frac { 1 }{ 2 } { w }^{ T }w\quad +\quad C\sum _{ i=1 }^{ m }{ { \xi  }_{ i } }  
\label{eqn:svm}
\end{equation}
\begin{equation}
s.t.\quad { y_i }.({ x_i }^{ T }w+b)\ge 1-{ \xi  }_{ i };\quad { { \xi  } }_{ i }\ge 0
\notag
\end{equation}

\begin{figure}[t]
  \centering
  \includegraphics[width=0.5\linewidth]{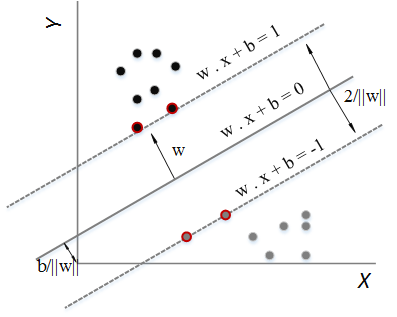}
   \caption{SVM with margin $w.x \pm b=1$.}
  \label{fig:svm}
\end{figure}

The above equation learns a linear boundary given by $w.x \pm b=1$, which separates the two classes with a margin of width $2/||w||$, as shown in Figure~\ref{fig:svm}. The trained SVM model, has an expected number of samples in the margin, due to its soft constraints. The margin density here is given by the ratio of samples which fall inside the margin of the SVM, as per Equation~\ref{eqn:mdsvm}. The signal function $S_{(w,b)}(x)$ checks if a given sample $x$ falls within the margin of the SVM, with parameters $w$ and $b$.

\begin{equation}
\begin{aligned}
{ MD }_{ SVM }\quad =\quad \frac { \sum { { S }_{ (w,b) }(x) }  }{ \left| X \right|  } ;\quad \forall x\in X
\\
where, \\ 
\quad { S }_{ (w,b) }(x)\quad =\quad \begin{cases} 1,\quad if\quad \left| w.x+b \right| \quad \le \quad 1 \\ 0,\quad otherwise \end{cases}
\end{aligned}
\label{eqn:mdsvm}
\end{equation}

The set of unlabeled samples is given by $X$ and the distance from the hyperplane is given by $|w.x+b|$. This distance is threshold by a $sign()$ function, to produce the final class label of +1 or -1. If the distance from the hyperplane, for a sample $x$, is within the margin ($\leq 1$), then the signal function S emits a 1; denoting that it contributes to the margin density. For other probabilistic classifiers, such as logistic regression, which return probability of class +1 or -1 as $p(y=+1|x)$ and $p(y=-1|x)$, the confidence is computed as $|p(y=+1|x) - p(y=-1|x)|$ and the threshold $\theta_{margin}$ is used to specify the cutoff for uncertain samples. Samples with confidence less than $\theta_{margin}$, contribute to the margin density.

\subsubsection{Classifiers without explicit margins}
\label{sec:computing_rs}

Classifiers such as decision trees \citep{quinlan1993c4} and K-nearest neighbors \citep{cover1967nearest} return discrete class labels and they do not have any intuitive notion of margin. These models are considered unstable \citep{dietterich2000ensemble} and they reduce the number of features necessary to build the models. In order to make them robust and to distribute weights across features, they are used with a feature bagging ensemble technique. Feature bagging improves generalization of unstable classifiers, by training multiple base models of the classifier on different subset of features, from the original D-dimensional data space \citep{bry}. Random subspace \citep{ahn2007classification,skurichina2002bagging} is an implementation strategy for feature bagging and is given by Algorithm~\ref{algo:rs}. The entire feature space of $D$ features is divided into $K$ randomly chosen subspaces, with $J$ features each, and a classifier is trained on each of these subspaces. The resulting ensemble $E$ has classifiers $C_{i}; i=1..K$, and majority voting is used to predict the final label $y$ for a given sample $x$. By employing random subspace ensemble, any base classifier type can be made robust and the margin density signal can be extracted from it.

In Algorithm~\ref{algo:rs}, the set of \textit{K} models in the ensemble \textit{D}, are trained on different views of the feature space and serve as a committee of experts with independent views on the prediction problem. An increased disagreement between the models is indicative of high uncertainty over a sample. The margin density MD for this type of models is computed by measuring the number of samples which have high uncertainty, as given by Equation~\ref{eqn:rs}. The Signal function $S_E(x)$, checks to see if the sample $x$ has certainty less than $\theta_{margin}$ (typically taken as 0.5). In the equation below, $p_E$ refers to the voted mean predicted class probabilities of the base estimators in the ensemble. 

\begin{equation}
\begin{aligned}
{ MD }_{ RS }=\frac { \sum { { S }_{ E }(x) }  }{ \left| X \right|}; \forall x\in X \\
where,\\
{ S }_{ E }(x)=\begin{cases} 1, if \quad \left| { p }_{ E }({ y= -1 }|x)-{ p }_{ E }({ y= +1 }|x) \right| \\ 
\quad\quad\quad\quad \le { \theta  }_{ margin } \\ 0, \quad otherwise \end{cases}
\end{aligned}
\label{eqn:rs}
\end{equation}

The set of unlabeled samples $X$ is collected and the ratio of samples which have critical uncertainty ($\leq \theta_{margin}$), is given as the margin density. The ensemble $E$ can be comprised of classifier of any type (even heterogeneous classifiers can be considered), making the margin density approach applicable irrespective of the choice of the classification algorithm used.

{\LinesNotNumbered
\begin{algorithm}[t]
\caption {Random Subspace Ensemble.}
\label{algo:rs}
\underline{\textit{\textbf{Training:}}} \\
 \For{ i= 1,2,3,...,K:}{
 
	Select J random features from all D features
	
	Construct $C_i$ in $X_J$ and add it to ensemble E
	
  }
\underline{\textit{\textbf{Classification:}}}\\
   Use majority voting to provide prediction on the input sample x. $y(x)=argmax_y(votes(y))$
\end{algorithm}}

\subsection{Change in Margin Density ($\Delta MD$) as an indicator of drift}
\label{sec:proofofconcept}

To understand the behavior of the margin density metric and its efficacy as an indicator of drift, it is evaluated here on a synthetic dataset, under different change scenarios. A change scenario is setup by generating an initial distribution of 500 samples, used for training a model, and then generating 500 additional samples from a changed distribution, for testing the model. The change in margin density ($\Delta MD$) is evaluated as the difference in margin densities of the training and test data: $\Delta MD=|MD_{Train}-MD_{Test}|$. By comparing $\Delta MD$ with changes in the training and testing error ($\Delta Err$), which is representative of a metric used by fully labeled drift detectors, the effectiveness of MD to detect true drifts is evaluated. Similarly, a comparison with traditional feature based unlabeled drift detection techniques is evaluated, by checking the Hellinger distance between the distributions ($\Delta HD$) \citep{ditzler2011hellinger}. Since $\Delta Err$ and $\Delta MD$ are within the range [0,1] and $\Delta HD$ has a range of [0,$\sqrt[]{2}$], the $\Delta HD$ values were normalized to [0,1] by dividing the values by $\sqrt[]{2}$ in all the following experiments. The margin density metric was evaluated for a linear kernel SVM and for a Random subspace ensemble with 2 orthogonal C4.5 decision trees. 

\begin{figure}[t]
\centering
\subfloat[A0 - Initial(SVM)]{\includegraphics[width=0.2\linewidth]{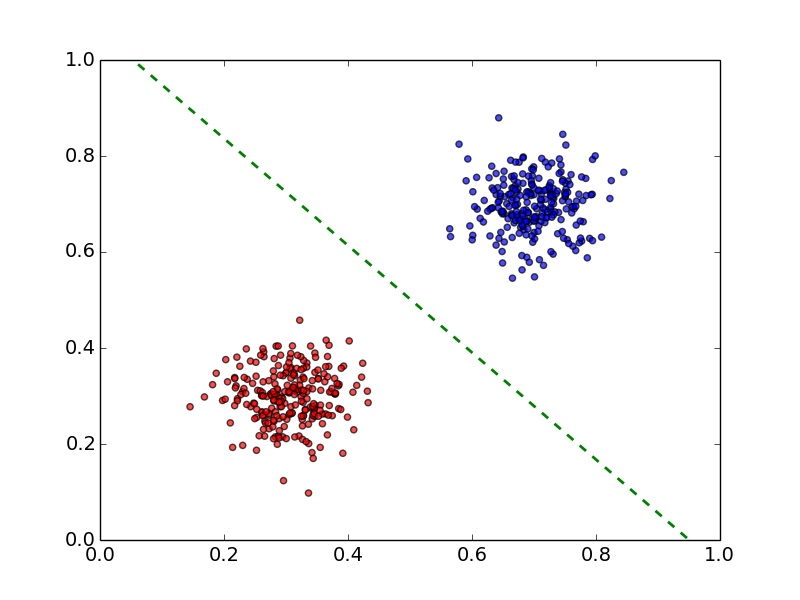}}
\subfloat[A0-A1]{\includegraphics[width=0.2\linewidth]{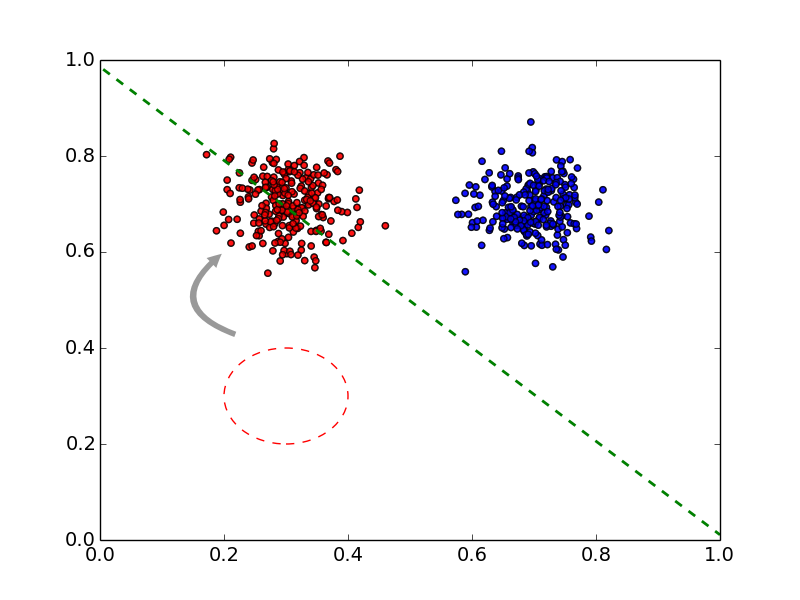}}
\subfloat[A0-A2]{\includegraphics[width=0.2\linewidth]{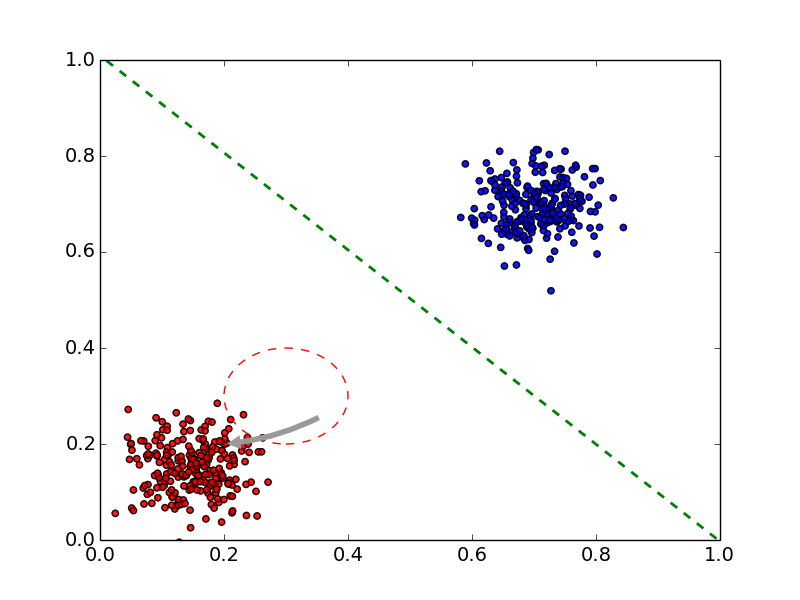}}
\subfloat[A0-A3]{\includegraphics[width=0.2\linewidth]{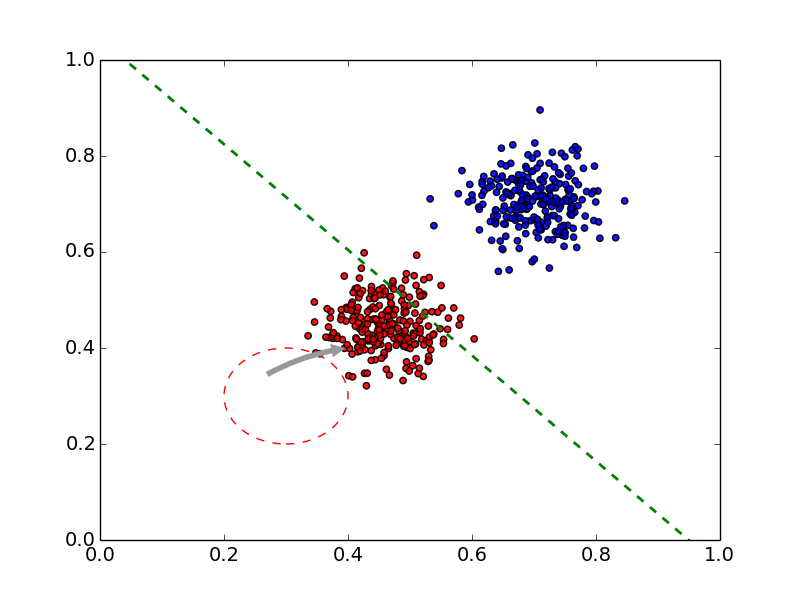}}
\subfloat[A0-A4]{\includegraphics[width=0.2\linewidth]{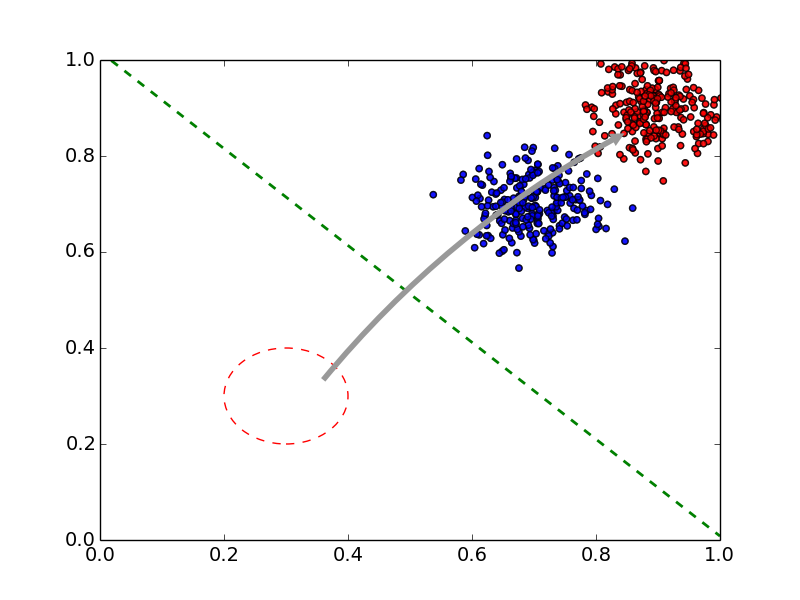}}
\\
\subfloat[A0 - Initial(RS)]{\includegraphics[width=0.2\linewidth]{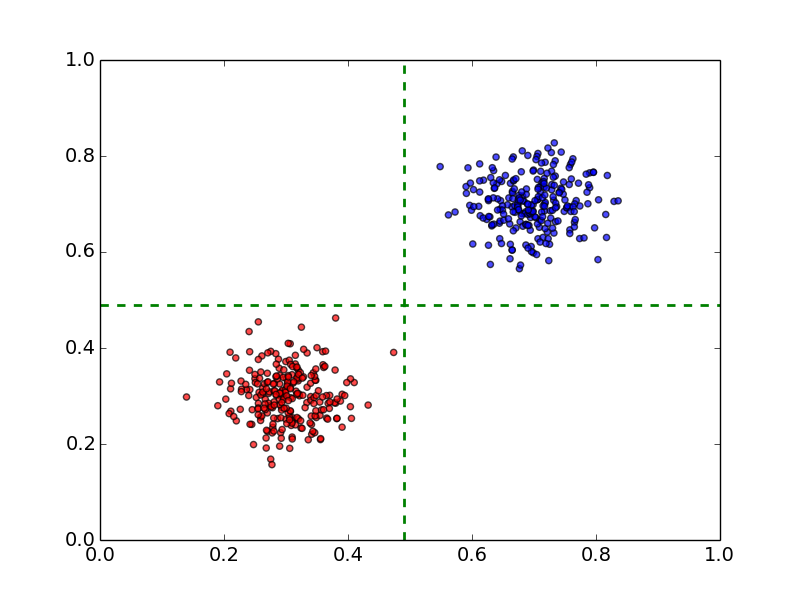}}
\subfloat[A0-A1]{\includegraphics[width=0.2\linewidth]{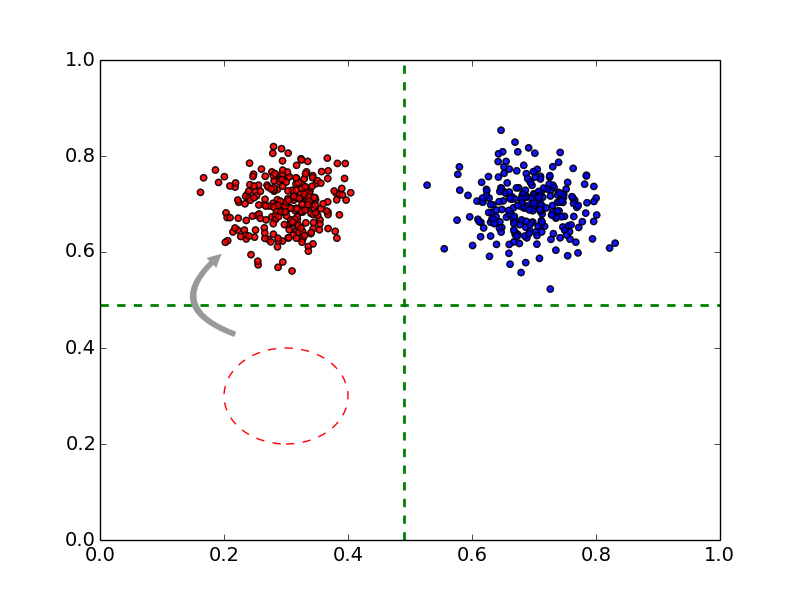}}
\subfloat[A0-A2]{\includegraphics[width=0.2\linewidth]{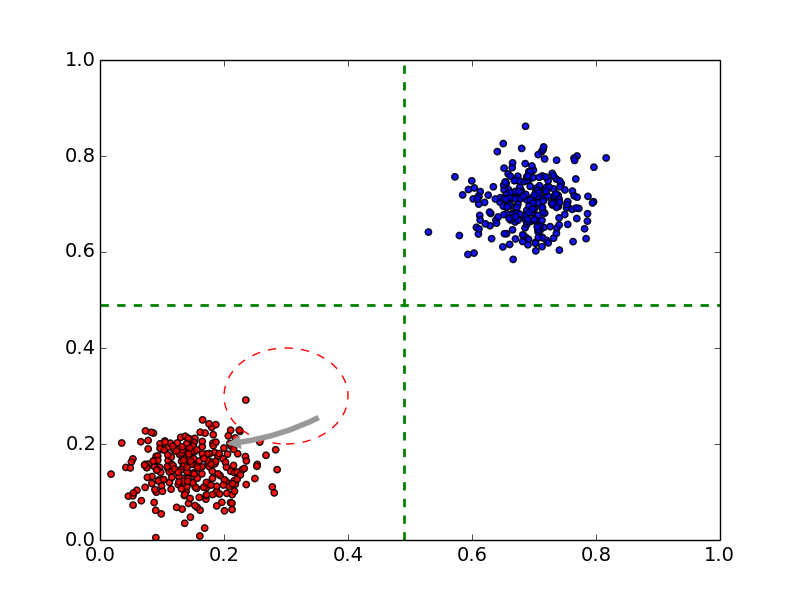}}
\subfloat[A0-A3]{\includegraphics[width=0.2\linewidth]{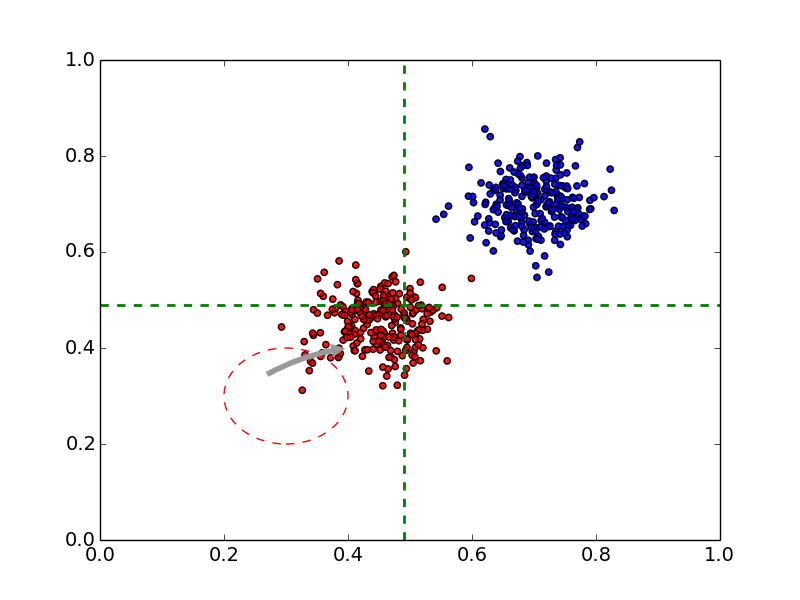}}
\subfloat[A0-A4]{\includegraphics[width=0.2\linewidth]{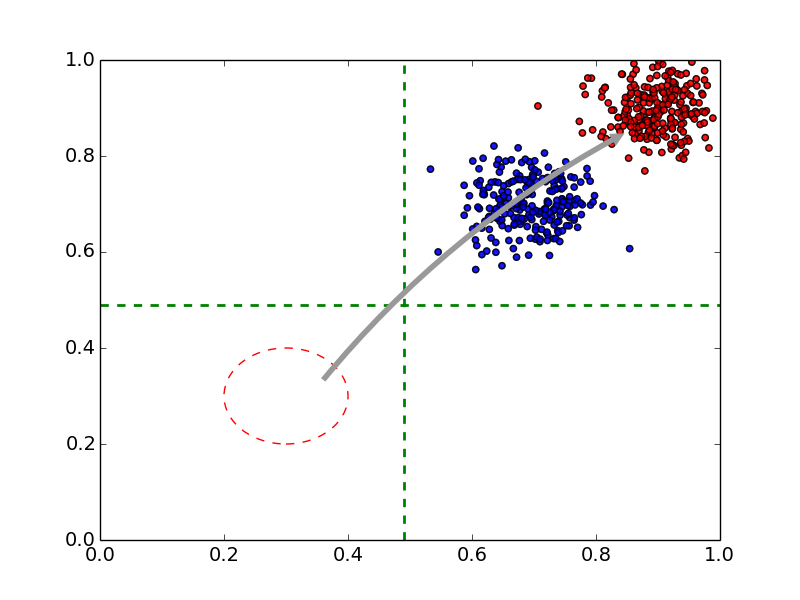}}
\caption{Drift Scenarios with SVM (top) and random subspace (RS) (bottom) on 2D synthetic dataset. A0 is the initial distribution. A1-A4 represent different drift scenarios.}
\label{fig:scenarios1}
\end{figure}

\begin{table}[t]
\centering
\caption{Results of change detection metrics $\Delta Err$, $\Delta MD$ and $\Delta HD$, on synthetic drifting scenarios.}
\label{tbl:syntheic}
\resizebox{0.45\textwidth}{!}{\begin{tabular}{|c|l|l|l|l|l|l|}
\hline
\textbf{Model-} & \multicolumn{3}{c|}{\textbf{SVM}} & \multicolumn{3}{c|}{\textbf{RS}} \\ \hline
\textbf{Drift} & \multicolumn{1}{c|}{\textbf{ΔErr}} & \multicolumn{1}{c|}{\textbf{ΔMD}} & \multicolumn{1}{c|}{\textbf{ΔHD}} & \multicolumn{1}{c|}{\textbf{ΔErr}} & \multicolumn{1}{c|}{\textbf{ΔMD}} & \multicolumn{1}{c|}{\textbf{ΔHD}} \\ \hline
\textit{A0-A1} & 0.19 & 0.43 & 0.33 & 0.5 & 0.5 & 0.33 \\ \hline
\textit{A0-A2} & 0 & -0.04 & 0.58 & 0 & 0 & 0.58 \\ \hline
\textit{A0-A3} & 0.05 & 0.42 & 0.58 & 0.1 & 0.17 & 0.58 \\ \hline
\textit{A0-A4} & 0.5 & -0.05 & 0.69 & 0.5 & 0 & 0.69 \\ \hline
\textit{B0-B1} & 0 & 0.02 & 0.27 & 0 & 0 & 0.27 \\ \hline
\textit{C0-C1} & 0.41 & -0.72 & 0.82 & 0.41 & -0.19 & 0.82 \\ \hline
\end{tabular}}
\end{table}

\begin{figure}[t]
\centering
\subfloat[B0 - Initial(RS)]{\includegraphics[width=0.45\linewidth]{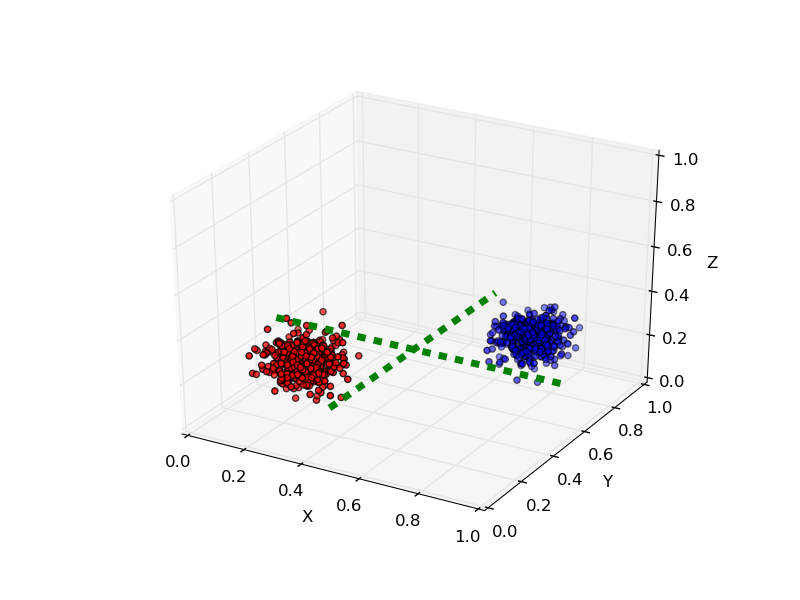}}
\subfloat[B0-B1]{\includegraphics[width=0.45\linewidth]{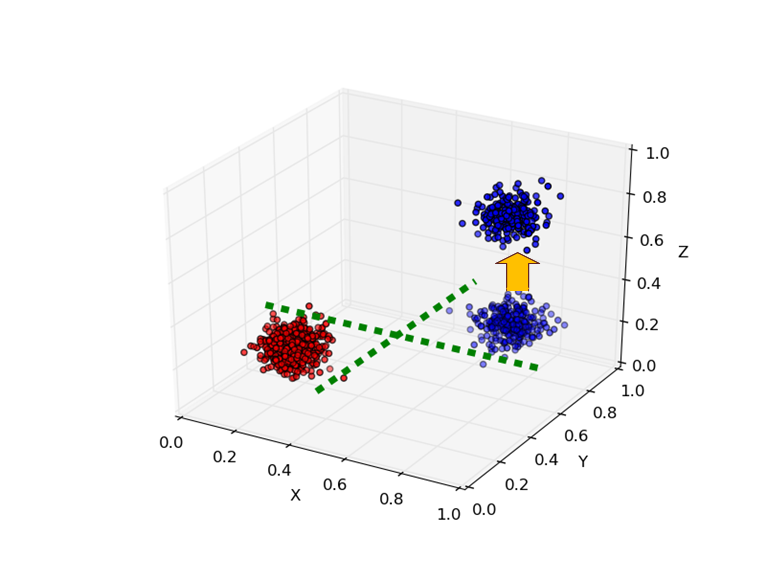}}
\caption{Drift Scenario in 3D synthetic dataset, change occurs along Z- dimension, which is irrelevant to the classification task. }
\label{fig:scenarios2}
\end{figure}

Experiments on a 2D synthetic dataset with two classes, using an SVM and a random subspace(RS) model are depicted in Figure~\ref{fig:scenarios1}. A0 represents the initial training distribution of the samples and A1-A4 represent 4 different change situations. The drift scenario A0-A1 represent changes which directly affects classification boundary, due to drift in one of the features. This causes the error to increase by 19\% for SVM and 50\% for RS, as shown in Table~\ref{tbl:syntheic}. Correspondingly, the $MD$ changes by an average of 0.47 and the $HD$ changes by 0.33. Since the Hellinger distance is computed from the unlabeled data, independently of the learned classifier, the $\Delta HD$ values for SVM and RS are same for any given scenario, as shown in Table~\ref{tbl:syntheic}. Change scenarios A0-A2 and A0-A3 represent shift of equal magnitude but opposite direction. In A0-A2, the shift is away from the margin, while in A0-A3 the shift is towards the margin, as shown in Figure~\ref{fig:scenarios1}c) and d) for SVM, and h) and i) for RS models. Both these scenarios result in the same change in $\Delta HD$ of 0.58, indicating shortcomings of traditional feature tracking approaches in differentiating false alarms from relevant changes. The $MD$ metric shows no change for the A0-A2 scenario but detects the A0-A3, consistent with the error tracking approach. By using a fixed margin and tracking its density, changes away from the margin are effectively ignored, as they rarely cause performance degradation. This property of the margin density approach makes it more resistant to false alarms, compared to other margin based methods \citep{dries2009adaptive,zliobaite2010change}. An extreme data distribution shift is seen in Figure~\ref{fig:scenarios1}e) and j), which is representative of a drastic drift affecting all features simultaneously; a situation rare in real world applications. This change occur away from the margin and goes unnoticed by the $\Delta MD$  metric. They are however, tracked by the $\Delta HD$ and $\Delta Err$ methods. These changes are rare in real world operational environments and can be more effectively caught by specialized novelty detection methods developed for such cases  \citep{masud2011classification, farid2013adaptive}. The Hellinger distance metric, in an attempt to provide completeness in drift detection, leads to excessive false alarms. This effect is exacerbated in high dimensional datasets, where there are many irrelevant features that do not contribute to the classification process. This is illustrated in Figure~\ref{fig:scenarios2}, where the Z-dimension is not useful to the prediction task. A drift in the Z- direction leads to a false alarm by $\Delta HD$, but is correctly ignored by $\Delta Err$ and $\Delta MD$, as seen for the entry B0-B1 in Table~\ref{tbl:syntheic}. 

\begin{figure}[t]
\centering
\subfloat[C0 - Initial(SVM)]{\includegraphics[width=0.35\linewidth]{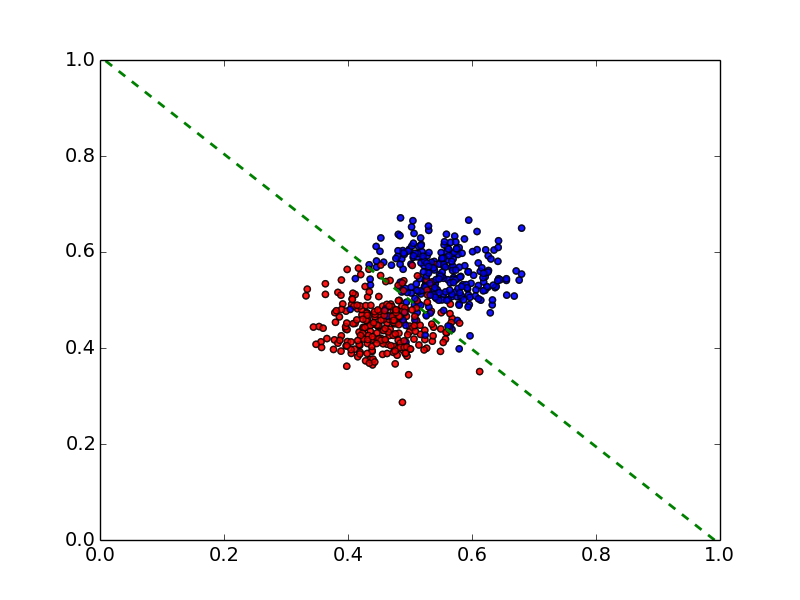}} \quad
\subfloat[C0-C1(SVM)]{\includegraphics[width=0.35\linewidth]{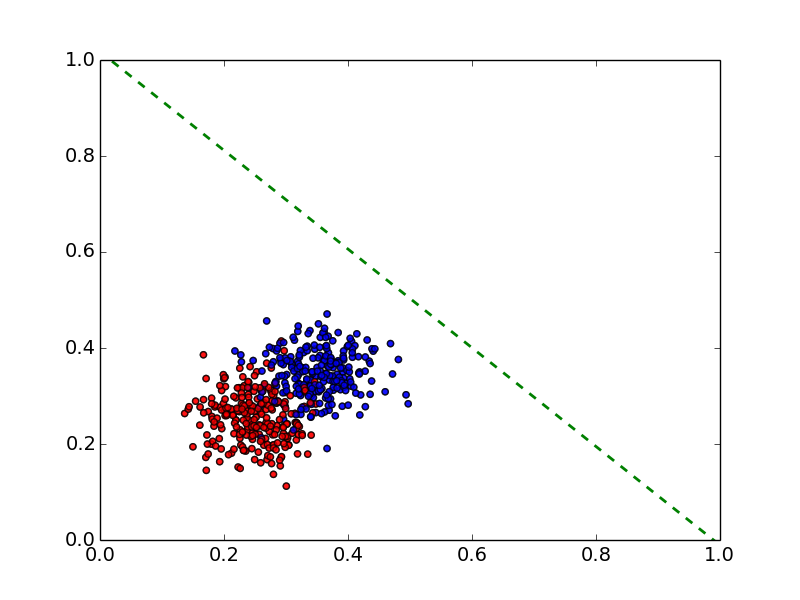}}
\\
\subfloat[C0 - Initial(RS)]{\includegraphics[width=0.35\linewidth]{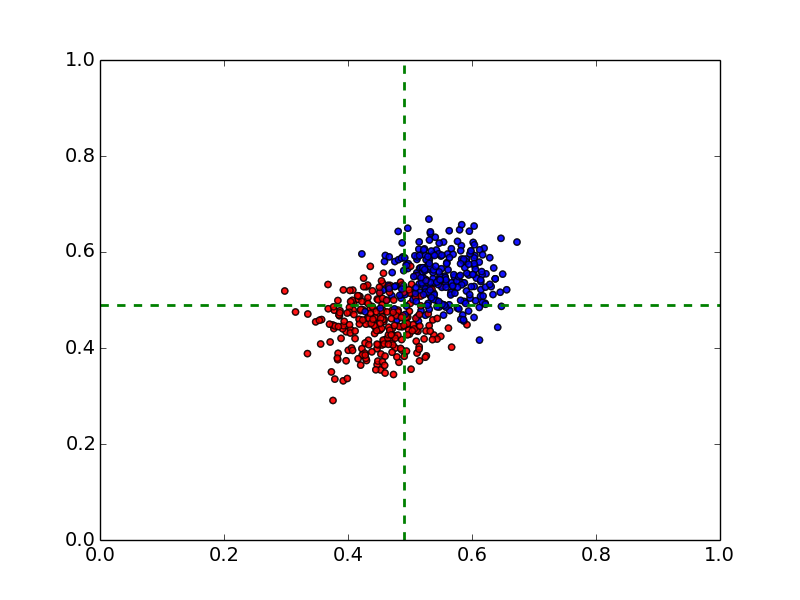}} \quad
\subfloat[C0-C1(RS)]{\includegraphics[width=0.35\linewidth]{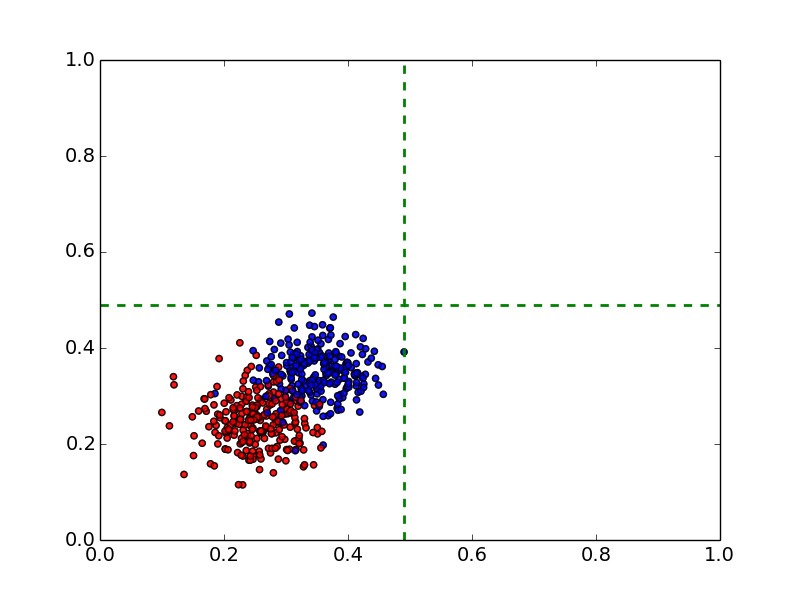}}
\caption{Drift scenario causing drop in margin density, with SVM model (top) and random subspace model (RS) (bottom). C0 is the initial distribution of samples. }
\label{fig:scenarios3}
\end{figure}

An additional scenario C0 is presented, to indicate the need for tracking both a drop as well as a rise in the margin density. The scenario C0-C1 as shown in Figure~\ref{fig:scenarios3}, is common in case of non linear or tightly packed class distributions, which causes the initial margin density to be high. This drift leads to a drop in the margin density, indicated by negative values in Table~\ref{tbl:syntheic}, indicating the need to track the absolute value of margin density change: $|\Delta MD|$. This is in contrast to the $\Delta Err$ and $\Delta HD$ metric, where only a spike in error rate or Hellinger distance is considered relevant to change detection. 

The analysis in this section indicates the ability to use the change in margin density ($\Delta MD$) as a signal for drift detection. The MD metric signals change when it is relevant to the classification process, while providing high robustness against stray changes. It can therefore be used as a surrogate to labeled error tracking approach, to provide reliable drift detection. Change in irrelevant features, in high dimensional spaces and in regions away from the classifier's margin, are effectively filtered. This effect was observed for both - classifiers with explicit margin and by using a random subspace ensemble for cases where margin is not explicit.

\section{The Margin Density Drift Detection (MD3) algorithm}
\label{sec:md3}

\begin{figure}[t]
  \centering
  \includegraphics[width=0.9\linewidth]{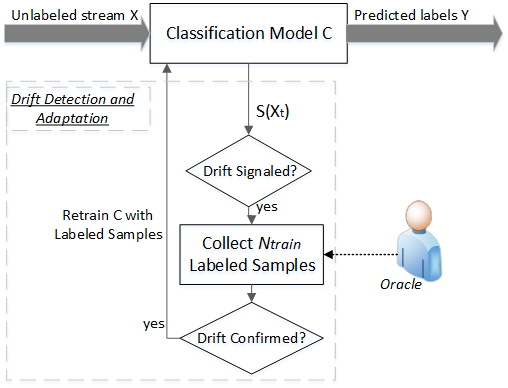}
   \caption{Overview of the MD3 methodology.}
  \label{fig:framework}
\end{figure}

The MD3 algorithm, being a streaming data algorithm, needs to operate with limited memory of past information, needs to continuously process data indefinitely over a single pass, and has to provide a quick response time. The change in margin density ($\Delta$MD) is used as a metric for detecting drift in a streaming environment. The incremental classification process continuously receives unlabeled samples \textit{X} and predicts their class labels \textit{Y}, based on the classification model \textit{C}, as shown in Figure~\ref{fig:framework}. At any given time \textit{t}, the signal function $S(X_t)$ computes if the sample $X_t$ lies within the margin of \textit{C}. This computation is performed using Equations~\ref{eqn:mdsvm} and \ref{eqn:rs}, based on the type of model used. This signal is used to update the expected margin density. A significant change in the margin density at time \textit{t} ($MD_t$) signals a change which requires further inspection. Following this, the next $N_{train}$ samples are requested to be labeled by an external Oracle. The Oracle can be any entity which can provide true labels for the unlabeled sample $X_t$, at a given cost. If the performance of \textit{C}, on the $N_{train}$ labeled samples, is found to have degraded, a drift is confirmed and the model is retrained using these labeled samples collected. In the MD3 approach, there is no need for continuous monitoring using labeled samples, as the drift detection process is unsupervised. Labeling is requested only when a drift is suspected, for confirmation and retraining. The MD metric reduces the need for frequent confirmation, owing to its robustness toward irrelevant changes, making the labeling requested essentially for the retraining phase only.

\begin{algorithm}[t]
\SetKwInOut{Input}{Input}
\SetKwInOut{Output}{Output}
 \Input{ Unlabeled stream X, Initially trained model \textit{C}, Reference distribution ($MD_{Ref}$, $\sigma_{Ref}$, $Acc_{Ref}$, $\sigma_{Acc}$). \textit{Parameters:} Sensitivity $\Theta$, Stream progression $\lambda=(N-1)/N$(where N is the chunk size), $N_{train}$(= N by default)}
 \Output{Predicted label stream Y}
 
 $MD_0$ = $MD_{Ref}$
 
 currently\_ drifting = False
 
 LabeledSamples = $\emptyset$
 
 \For{ t= 1,2,3,...:}{
 
 	Compute margin inclusion signal- 
 	$	S(x=X_{ t })=\begin{cases} 1,\quad if\quad { X }_{ t }\quad in\quad margin \\ 0,\quad otherwise \end{cases}
	$
		
	Update $MD_t$ = $\lambda$ * $MD_{t-1}+(1-\lambda)*S(X_t)$
	
	\If{$|MD_t-MD_{Ref}|\textgreater \Theta$ * $\sigma_{Ref}$}
	{ // \textbf{\textit{Drift Suspected}}
	
		currently\_drifting = True
		
		LabeledSamples$\leftarrow$Collect $N_{train}$ labeled samples by querying Oracle
	}
	\If{currently\_ drifting and $|LabeledSamples|$== $N_{train}$}
	{
		//\textit{Enough labeled samples to make decision}
		
		\If{$Acc_{Ref}$-($Acc_{LabeledSamples}$)$\textgreater \Theta$  * $\sigma_{Acc}$}
		{// \textbf{ \textit{Drift Confirmed}}
		
			Retrain C with LabeledSamples 
		}
		Update Reference distribution ($MD_{Ref}$, $\sigma_{Ref}$, $Acc_{Ref}$, $\sigma_{Ref}$)
		
		currently\_ drifting = False
	}
	}
\caption {The MD3 algorithm.}
\label{algo:md3}
\end{algorithm}

The MD3 algorithm (Algorithm~\ref{algo:md3}), begins with an initial trained classifier \textit{C}, which is obtained by learning from the initial labeled training dataset, before the model is made online. From this initial training dataset, a reference distribution summarizing margin and performance characteristics of the dataset, is learned. This reference distribution comprises of the expected margin density - $MD_{Ref}$, the acceptable deviation of the margin density metric- $\sigma_{Ref}$, expected accuracy on the training dataset - $Acc_{Ref}$ and its deviation  $\sigma_{Acc}$. These values are learned from the training dataset by using the \textit{K}-fold cross validation technique, commonly used for evaluating classifiers \citep{kohavi1995study}. In the cross validation method, the entire dataset is sequentially divided into \textit{K} bands of samples. In the first iteration, the first \textit{K}-1 bands are used as a training dataset to learn model \textit{C} and then the $K^{th}$ band is used to test the model. The process is repeated \textit{K} times, where each band functions as the test set exactly once. Accuracy and Margin density values from the \textit{K} test sets are considered results of random experimentation. The average values and standard deviation, of the test accuracy and margin density, over the \textit{K} iterations is used to form the reference distribution. Cross validation allows to create a population of the metric values, to better estimate their expected values and acceptable deviation. These values are then used to signal change based on the desired level of sensitivity, given by parameter $\theta$. Change is signaled when the margin density at a time t, given by $MD_t$, deviates by more than $\theta$ standard deviations from the reference margin density  value $MD_{Ref}$, as given by (\ref{eqn:changemd}) (Line 7). The same sensitivity parameter is used to detect significant drop in performance, for the obtained labeled samples, from the reference accuracy values, as per (\ref{eqn:changeacc}) (Line 14). 

\begin{equation}
\label{eqn:changemd}
\begin{split}
if \quad |MD_t-MD_{Ref}| > \theta * \sigma_{Ref} \\
\Rightarrow Drift\quad suspected
\end{split}
\end{equation}
\begin{equation}
\label{eqn:changeacc}
\begin{split}
if \quad (Acc_{Ref}-Acc_{LabeledSamples}) > \theta * \sigma_{Acc} \\
 \Rightarrow Drift\quad confirmed
\end{split}
\end{equation}

Here, \textit{LabeledSamples} is the set of $N_{train}$ samples, which were requested to be labeled once a significant drift is suspected by (\ref{eqn:changemd}). A drop in accuracy confirms that the change is indeed a result of concept drift and that model retraining is necessary, to update the classifier \textit{C}. Once retraining is performed, a new reference distribution ($MD_{Ref}$, $\sigma_{Ref}$, $Acc_{Ref}$, $\sigma_{Acc}$) is learned from the \textit{LabeledSamples} set, based on the \textit{K}-fold cross validation technique described above. By allowing users to specify the intuitive parameter of sensitivity, suggested to be picked in the range of [0,3], the entire change detection process is made flexible to be used in different streaming environments. A larger value can be set if frequent signaling is not desired, alternatively a lower value could be used for critical applications where small changes could be harmful, if undetected.

The drift detection process, set around tracking the margin density signal MD, is made incremental by using the moving average formulation of Equation~\ref{eqn:mdavg} (Line 6). Here, the margin density at a time \textit{t}, given by $MD_t$, is computed incrementally by using a forgetting factor $\lambda$ on - $MD_{t-1}$, and combining it with the signal function $S(X_t)$, which indicates if the current sample $X_t$ falls within the margin of the classifier \textit{C}. 

\begin{equation}
\label{eqn:mdavg}
MD_t = \lambda*MD_{t-1} +(1-\lambda)*S(X_t)
\end{equation}

The parameter $\lambda$ is the forgetting actor for the stream and it can be computed by specifying the \textit{chunk of influence} parameter-\textit{N}. The $\lambda$ is computed as $\lambda=(N-1)/N$. This formulation makes it applicable as a stream monitoring system, by making incremental updates to the margin density metric. It should be noted that the incremental formulation is specified for the drift \textit{monitoring} process only. This is irrespective of the stream classification algorithm used, which could process data either - incrementally, by chunk or by using a sliding window. Separating the detection and classification schemes makes the MD metric more general in its implementation, to be used in different classification setups, and also enables the controlled testing of its efficacy in detecting meaningful concept drifts.

\section{Experimental evaluation}
\label{sec:experiment}

This section presents experimental analysis and results of the proposed MD3 approach on two sets of experiments: Section~\ref{sec:driftinduced} presents results on drift induced datasets, to better understand drift detection characteristics of the framework in a controlled environment; Experiments with real world drifting data are presented in Section~\ref{sec:cdd} and Section~\ref{sec:benchmark}, to demonstrate its practicality. Experimental comparisons with a fully labeled drift detection technique -the \textit{AccTr} approach based on EWMA \citep{ross2012exponentially}, and the unlabeled drift detection approach of- \textit{HDDDM} \citep{ditzler2011hellinger}, were performed. Two variants of the MD3 approach are evaluated- \textit{MD3-SVM} uses a linear kernel SVM as the base model, while \textit{MD3-RS }uses a random subspace implementation of the margin density approach. Details about experimental methods and setup are presented in Section~\ref{sec:experimental_methods}. Effects of varying the margin width parameter $\theta_{margin}$ and that of varying the detection model, are presented in Section~\ref{sec:modelsmargins}. 

\subsection{Experimental methods and setup}
\label{sec:experimental_methods}

\subsubsection{Experimental methods used for comparative analysis}
\label{sec:methods_comparative}

The drift detection on data streams are evaluated and compared using the following techniques. 

\begin{figure}[t]
  \centering
  \includegraphics[width=0.9\linewidth]{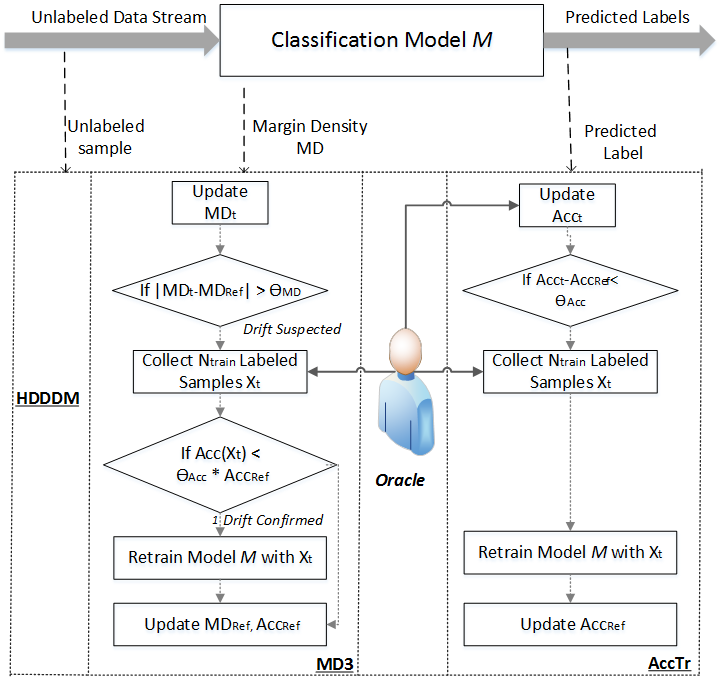}
   \caption{Unlabeled (HDDDM), Fully Labeled (AccTr) and the Margin Denstiy(MD3) drift detection techniques, showing portion of stream that they track.}
  \label{fig:comparison}
\end{figure}

\begin{itemize}
\item \textit{Static baseline model (\textbf{NoChange})}: This approach assumes that data is static with no drift over time. As such, no change is signaled, model is never updated and no labeling is requested. This is the lower baseline and any approach should atleast be better than a NoChange approach.

\item \textit{Fully labeled Accuracy Tracking (\textbf{AccTr})}: This model forms the upper baseline for the drift handling mechanisms. All data is assumed to be labeled and the explicit tracking of accuracy is used to signal change. An unlabeled drift detection mechanism is effective if its performance is close to the AccTr approach. The AccTr approach is illustrated in Figure~\ref{fig:comparison}, where every predicted sample's correct label is requested from an Oracle and the accuracy is checked to see if it has significantly deviated from the training accuracy. The accuracy is tracked incrementally by using the EWMA \citep{ross2012exponentially} formulation of change tracking as given by Equation~\ref{eqn:ewma}. 

\item\textit{MD3 using SVM model (\textbf{MD3-SVM})}: This approach uses the SVM based implementation of MD3. A linear kernel SVM with hinge loss is used. Margin density is computed by tracking number of samples in the classifier's margin. 

\item\textit{MD3 using random subspace(RS) model (\textbf{MD3-RS})}: The proposed random subspace drift detection technique of MD3 is used. The ensemble comprises of 20 C4.5 decision trees, each one having 50\% of the features randomly picked from the feature space. Margin density is given by number of samples in regions of high uncertainty (high disagreement), of the ensemble. The threshold for critical uncertainty was chosen as 0.5, and samples with confidence less than 0.5 are considered to be in the margin. 

\item\textit{Hellinger Distance Drift Detection Methodology (\textbf{HDDDM})}: The approach obtained from   \citep{ditzler2011hellinger}, is representative of traditional unlabeled feature tracking approaches found in literature, which track changes to feature space. In particular, the HDDDM approach tracks the average Hellinger distance of all samples within two distributions and signals change when the distance increases beyond a threshold. Hellinger distance is a popular metric in streaming data research, and a comparison using this will enable us to highlight the fundamental differences between the unlabeled approaches and the MD3 approach.
\end{itemize}

These methods provide representations of the main different paradigms of drift detection: Explicit detector(labeled) and Implicit detectors(unlabeled), as presented in Section~\ref{sec:review}. With a comparative analysis of these methods, we aim to will highlight the efficacy of the MD3 approach and its place in the literature on drift detection techniques. 

\subsubsection{Experimental setup}
\label{sec:setup}

To ensure that bias due to the underlying classification process does not affect our analysis of the detection scheme, all approaches were implemented in an incremental manner using the moving average formulation of Equation~\ref{eqn:metricavg}. 

\begin{equation}
\label{eqn:metricavg}
Metric_t = \lambda*Metric_{t-1} +(1-\lambda)*S
\end{equation}

The metric at time \textit{t} depends on the signal function \textit{S(.)}, which is defined based on the detection method used. The AccTr approach uses error as a signal. If predicted label is different from the correct label, the signal \textit{S} is 0, otherwise it is 1. For the MD3 approaches, the signal is obtained by the margin inclusion test as given by Equation~\ref{eqn:mdsvm} for MD3-SVM and Equation~\ref{eqn:rs} for the MD3-RS methods. The HDDDM approach is not incremental by nature. It is a chunk based approach, as computing histograms of data needs an entire chunk of data. We modified this approach, such that a chunk is defined incrementally, sliding at a rate of one sample. For a given time \textit{t}, the chunk comprises of $t-N$ latest samples, where \textit{N} is the chunk size. The $\lambda$ forgetting factor in Equation~\ref{eqn:metricavg} is taken as $(N-1)/N$ for AccTr and MD3 approaches, to ensure equivalence in the drift detection evaluation as compared to the HDDDM approach. 

The initial 15\% of the stream is assumed to be labeled. This forms the initial training set from which the classifier model \textit{C} is learned, along with the reference distribution metrics (expected metric and acceptable deviation). The reference distribution was obtained via 5-fold cross validation on the training data, as described in Section~\ref{sec:md3}, for the AccTr and MD3 approaches.  For the HDDDM approach, the reference distribution and acceptable deviation was obtained by using a sliding window of \textit{3*N} unlabeled samples with a slide rate of \textit{N/3} samples. The Hellinger distance between the subsequent chunks in this sliding window formed the population for learning the expected Hellinger distance and its standard deviation. A sensitivity of $\theta$=2 (from the suggested range of [0,3]) was chosen to balance robustness with reactivity, for all the drift detection methods.

All experiments were performed using Python 2.7 ad scikit-learn machine learning library \citep{scikit-learn}. Support Vector Machine with a linear kernel and regularization constant C = 1.0 was chosen as the prediction model for all the experiments. In order to ensure that differences between detection techniques MD3-SVM and MD3-RS are not a result of the different training capabilities of the SVM and RS classifiers, the task of prediction and detection were separated for the  MD3-RS approach. The training dataset was used to train two models: a linear SVM and a RS model. While, the SVM was used to provide prediction as the online classifier \textit{C} (Figure~\ref{fig:framework}), the RS model was used solely for the purpose of detection of drift and for triggering retraining of \textit{C}. This setup enables us to analyze the drift detection properties of the two approaches in a controlled environment, by blocking out their different prediction behavior.

\subsection{Experiments on drift induced datasets}
\label{sec:driftinduced}

This section presents experimental evaluation of the MD3 approach on static datasets, which were induced with concept drift in a controlled manner. By controlling the location and nature of drifts in these datasets, a better understanding of the drift detection capabilities of the different approaches, in a real world setting, is obtained. Six datasets were chosen from the UCI machine learning library \citep{Lichman:2013}, and they were pre-processed to have only numeric and binary values, normalized in the range of [0,1]. The multi-class datasets were reduced to a binary class problems and the data instances were shuffled randomly, to remove any unintended concept drifts already in the data. The characteristics of the datasets is shown in Table~\ref{tbl:induced_data}. The chunk size parameter \textit{N}, shown in Table~\ref{tbl:induced_data}, is used to process the stream based on the number of instances present in the datasets. The drift induction process is explained next, followed by experimental results and analysis on the drift induced data.

\begin{table}[t]
\centering
\caption{Characteristics of datasets chosen for drift induction experiments. }
\label{tbl:induced_data}
\resizebox{0.45\textwidth}{!}{\begin{tabular}{|l|l|l|l|}
\hline
Dataset & \#Instances & \#Attributes & Chunksize N \\ \hline
Digits08 & 1499 & 16 & 150 \\ \hline
Digits17 & 1557 & 16 & 150 \\ \hline
Musk & 6598 & 166 & 500 \\ \hline
Wine & 6497 & 12 & 500 \\ \hline
Bank & 45211 & 48 & 2500 \\ \hline
Adult & 48842 & 65 & 2500 \\ \hline
\end{tabular}}
\end{table}

\subsubsection{Inducing concept drift in static datasets}
\label{sec:induction}

The drift induction process of \citep{zliobaite2010change} provides a way to include a single concept drift in static datasets, at a particular location in the data stream. This allows for controlled drift analysis, while at the same time retaining properties of the real world applications, from which the the dataset is derived. The dataset is first shuffled to remove any unwanted concept drift and to  prepare it for the drift induction process. The drift induction process of \citep{zliobaite2010change} induces feature drift in the dataset after a point in the stream, called the \textit{ChangePoint}. Drift is induced by randomly picking a subset of the features and rotating their values, for a particular class. For example, if the feature (1,5,7) are picked for class label 0, after the \textit{ChangePoint} the instances belonging to class 0 have features (1,5,7) shuffled as (7,1,5). This basic approach ensures that feature drifts are induced while also maintaining the original data properties of the dataset. This approach is however dependent on the features selected for rotation and it provides erratic results if the 'right' set of features are not selected. 

Our drift induction approach, proposed here, extends the basic idea of \citep{zliobaite2010change}, and allows for greater control over the nature of change. Instead of randomly picking a set of features to be rotated, we pick features based on their importance to the classification task. This is done by ranking the features, based on their information gain metric \citep{duch2004comparison}, and then selecting features form the top or bottom of the list based on the nature of change desired. Two sets of experiments are performed here: a) The \textit{Detectability} experiments, which choose the top 25\% of features from the ranked list, and b) The \textit{False alarm} experiments, which choose the bottom 25\%. This was done to test the detection capabilities and robustness to irrelevant changes, respectively. The top 25\% of the ranked features have a high impact on the classification task, as these are ranked based on their information content, and modifying these features results in model degradation, which is necessary to be detected and fixed. Modifying the bottom 25\% of the features has less impact on the classification process and results in false alarms, which should be ignored by the detectors.

\begin{table}[t]
\centering
\caption{Effects of shuffling the top 25\% and the bottom 25\% feature, on the test accuracy.}
\label{tbl:induced_setup}
\resizebox{0.48\textwidth}{!}{\begin{tabular}{|l|l|l|l|l|}
\hline
\multicolumn{1}{|c|}{Dataset} & \multicolumn{1}{c|}{Train} & \multicolumn{1}{c|}{\begin{tabular}[c]{@{}c@{}}Test-\\ Original\end{tabular}} & \multicolumn{1}{c|}{\begin{tabular}[c]{@{}c@{}}Test- \\ Top 25\%\end{tabular}} & \multicolumn{1}{c|}{\begin{tabular}[c]{@{}c@{}}Test- \\ Bottom 25\%\end{tabular}} \\ \hline
Digits08 & 97.5 & 97.1 & 77.6 & 95.3 \\ \hline
Digits17 & 99.5 & 99.6 & 60.1 & 99.9 \\ \hline
Musk & 93.9 & 91.9 & 82.7 & 90.1 \\ \hline
Wine & 100 & 100 & 67.6 & 100 \\ \hline
Bank & 83.3 & 83.2 & 56.2 & 84.7 \\ \hline
Adult & 85 & 85.2 & 58.6 & 85.4 \\ \hline
\end{tabular}}
\end{table}

The effect of changing the top 25\% and bottom 25\% of the features on the 6 UCI datasets is shown in Table~\ref{tbl:induced_setup}. In all the datasets, the \textit{ChangePoint} was induced at 50\% of the stream. A model is trained on the data before the \textit{ChangePoint}, and tested on the samples after this point. The similarity in accuracy of the model on the training and original test set (before induction), indicates an initial static dataset. For the \textit{Detectability} experiments, the top 25\% of the features are rotated and this results in a significant drop in the test accuracy, after \textit{ChangePoint}, as seen in Table~\ref{tbl:induced_setup}. This indicates true drifts, which need to be detected by a drift detection algorithm. Rotating the bottom 25\% of the features, in the \textit{False alarm} experiments, does not show any significant drop in test accuracy. Although the same number of features are rotated in both cases, features have different levels of relevance, when it comes to the classification task. We perform our experimental analysis on the top 25\% and bottom 25\% datasets, to analyze behavior of the algorithms under different change conditions.

\subsubsection{Experimental results on drift induced datasets}
\label{sec:results_induced}

\begin{table*}[t]
\centering
\caption{Detection results on drift induced datasets for the \textit{Detectabiltiy} experiments and the \textit{False alarm} experiments.}
\label{tbl:induced_results}
\resizebox{0.7\textwidth}{!}{\begin{tabular}{|l|l|l|c|c|c|}
\hline
\multicolumn{1}{|c|}{\multirow{2}{*}{Dataset}} & \multicolumn{1}{c|}{\multirow{2}{*}{Methodology}} & \multicolumn{3}{c|}{\begin{tabular}[c]{@{}c@{}}Top25\%-\\ (\textit{Detectability} Expts)\end{tabular}} & \multicolumn{1}{c|}{\begin{tabular}[c]{@{}c@{}}Bottom25\%-\\ (\textit{False alarm} Expts)\end{tabular}} \\ \cline{3-6} 
\multicolumn{1}{|c|}{} & \multicolumn{1}{c|}{} & \multicolumn{1}{c|}{Accuracy} & \multicolumn{1}{c|}{Drifts detected} & \multicolumn{1}{c|}{False Alarms} & \multicolumn{1}{c|}{Drifts detected} \\ \hline
\multirow{5}{*}{Digits08} & NoChange & 86.4 & 0 & 0 & 0 \\ \cline{2-6} 
 & AccTr & 94.4 & 1 & 0 & 0 \\ \cline{2-6} 
 & MD3-SVM & 93.8 & 1 & 0 & 0 \\ \cline{2-6} 
 & MD3-RS & 94.4 & 1 & 0 & 0 \\ \cline{2-6} 
 & HDDDM & 93.5 & 2 & 1 & 1 \\ \hline
\multirow{5}{*}{Digits17} & NoChange & 71.9 & 0 & 0 & 0 \\ \cline{2-6} 
 & AccTr & 94.2 & 1 & 0 & 0 \\ \cline{2-6} 
 & MD3-SVM & 89 & 1 & 0 & 0 \\ \cline{2-6} 
 & MD3-RS & 93.2 & 1 & 0 & 0 \\ \cline{2-6} 
 & HDDDM & 88.7 & 2 & 1 & 3 \\ \hline
\multirow{5}{*}{Musk} & NoChange & 87 & 0 & 0 & 0 \\ \cline{2-6} 
 & AccTr & 94.1 & 1 & 0 & 0 \\ \cline{2-6} 
 & MD3-SVM & 94 & 2 & 1 & 0 \\ \cline{2-6} 
 & MD3-RS & 94.8 & 2 & 1 & 0 \\ \cline{2-6} 
 & HDDDM & 94.3 & 1 & 0 & 1 \\ \hline
\multirow{5}{*}{Wine} & NoChange & 80.1 & 0 & 0 & 0 \\ \cline{2-6} 
 & AccTr & 96.9 & 1 & 0 & 0 \\ \cline{2-6} 
 & MD3-SVM & 96.9 & 1 & 0 & 0 \\ \cline{2-6} 
 & MD3-RS & 94.9 & 1 & 0 & 0 \\ \cline{2-6} 
 & HDDDM & 96.9 & 3 & 2 & 1 \\ \hline
\multirow{5}{*}{Bank} & NoChange & 67.5 & 0 & 0 & 0 \\ \cline{2-6} 
 & AccTr & 89.4 & 1 & 0 & 0 \\ \cline{2-6} 
 & MD3-SVM & 89.4 & 1 & 0 & 0 \\ \cline{2-6} 
 & MD3-RS & 85.3 & 1 & 0 & 0 \\ \cline{2-6} 
 & HDDDM & 89.6 & 1 & 0 & 3 \\ \hline
\multirow{5}{*}{Adult} & NoChange & 67 & 0 & 0 & 0 \\ \cline{2-6} 
 & AccTr & 87.9 & 1 & 0 & 0 \\ \cline{2-6} 
 & MD3-SVM & 87.9 & 1 & 0 & 0 \\ \cline{2-6} 
 & MD3-RS & 87.9 & 1 & 0 & 0 \\ \cline{2-6} 
 & HDDDM & 88.2 & 3 & 2 & 3 \\ \hline
\end{tabular}}
\end{table*}

\begin{figure*}[ht]
\centering
\subfloat[Digits08]{\includegraphics[width=0.3\linewidth]{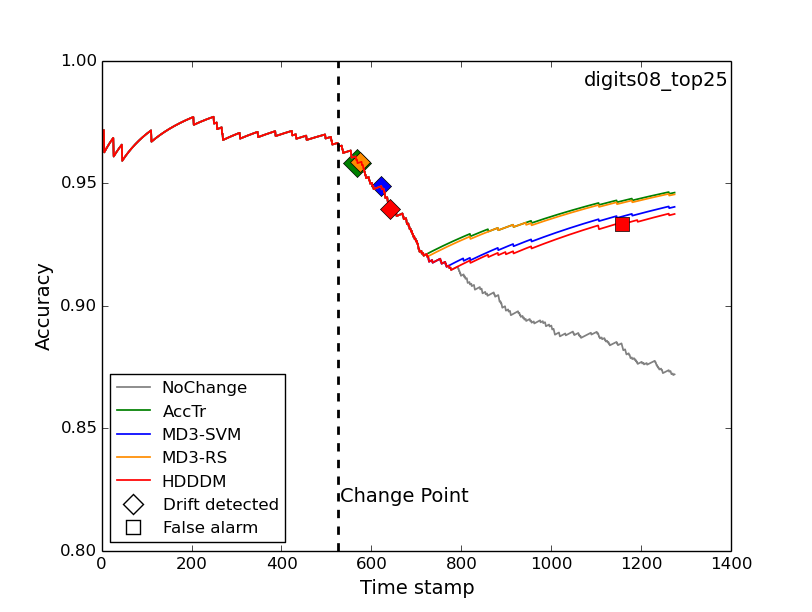}}
\subfloat[Digits17]{\includegraphics[width=0.3\linewidth]{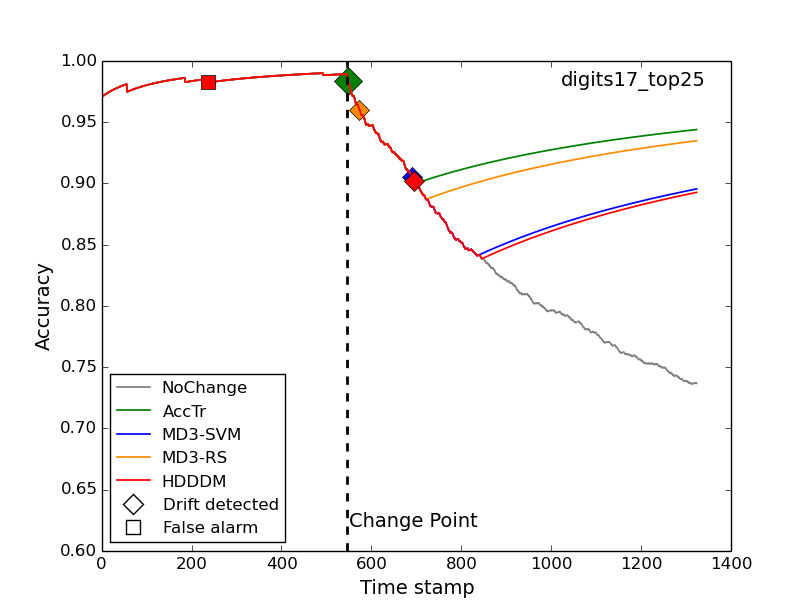}}
\subfloat[Musk]{\includegraphics[width=0.3\linewidth]{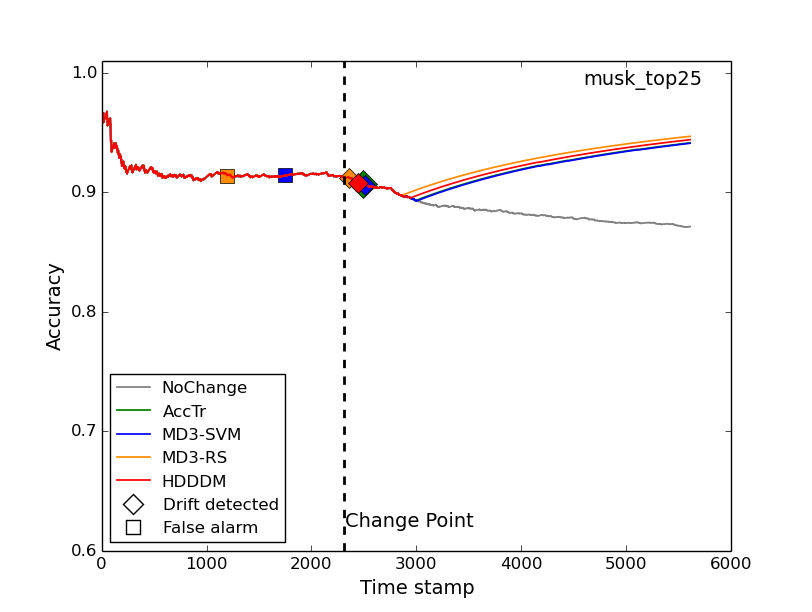}} \\
\subfloat[Wine]{\includegraphics[width=0.3\linewidth]{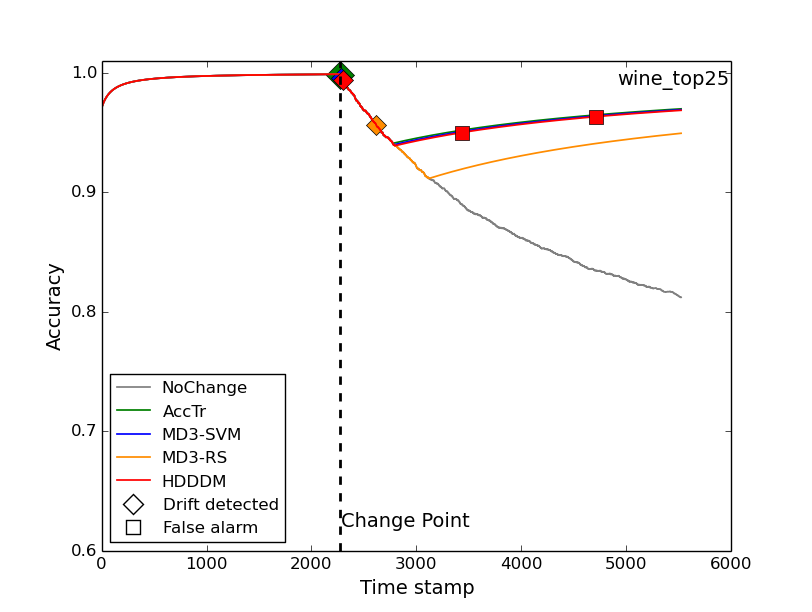}}
\subfloat[Bank]{\includegraphics[width=0.3\linewidth]{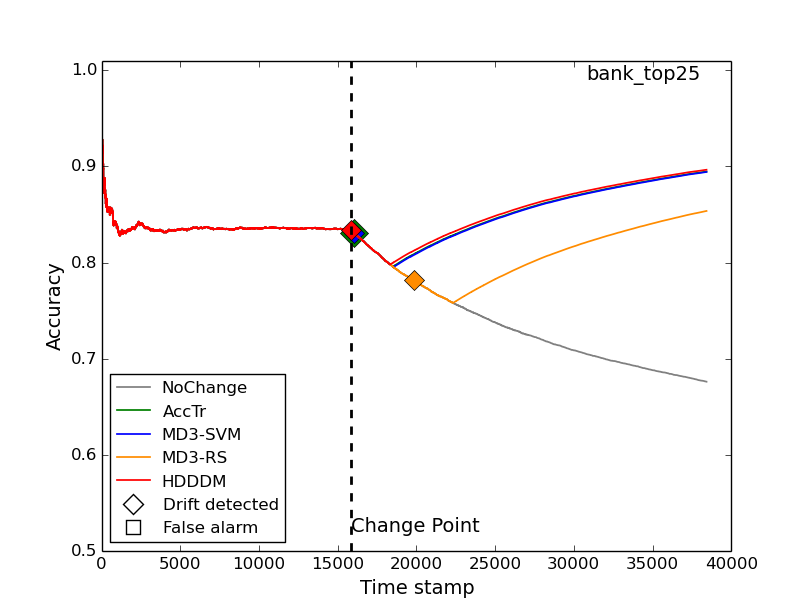}}
\subfloat[Adult]{\includegraphics[width=0.3\linewidth]{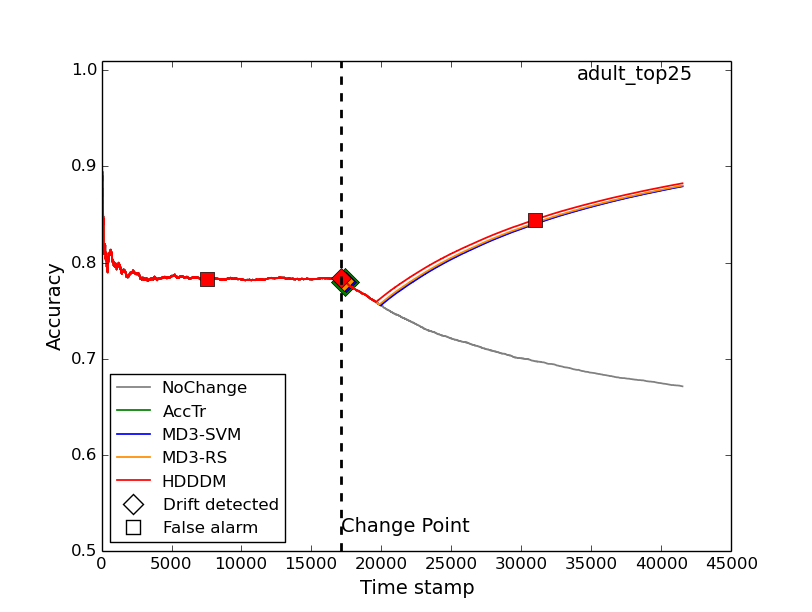}}
\caption{Accuracy over time for the NoChange(gray), AccTr(green), MD3-SVM(blue), MD3-RS(orange) and the HDDDM(red) approach on the \textit{Detectability} experiments. True drifts detected are shown as diamonds and squares represent false alarms. }
\label{fig:dd}
\end{figure*}

\begin{figure*}[ht]
\centering
\subfloat[Digits08]{\includegraphics[width=0.3\linewidth]{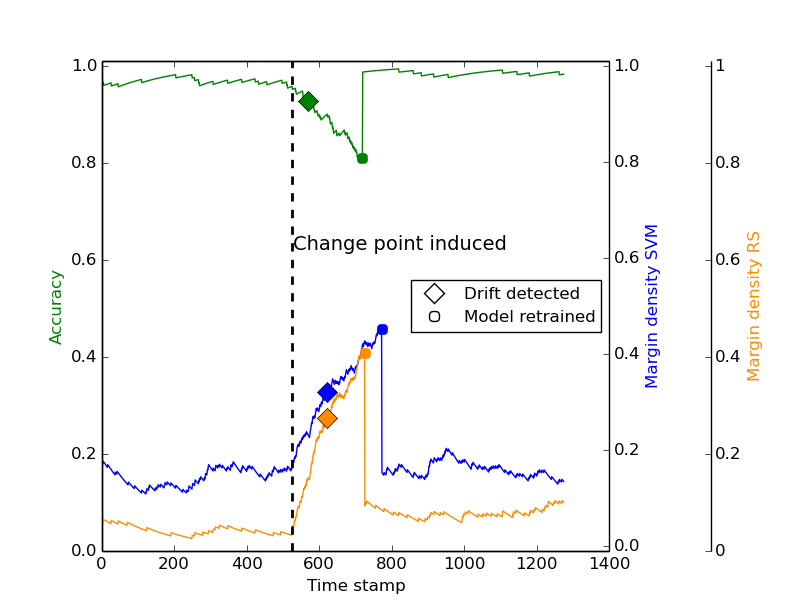}}
\subfloat[Digits17]{\includegraphics[width=0.3\linewidth]{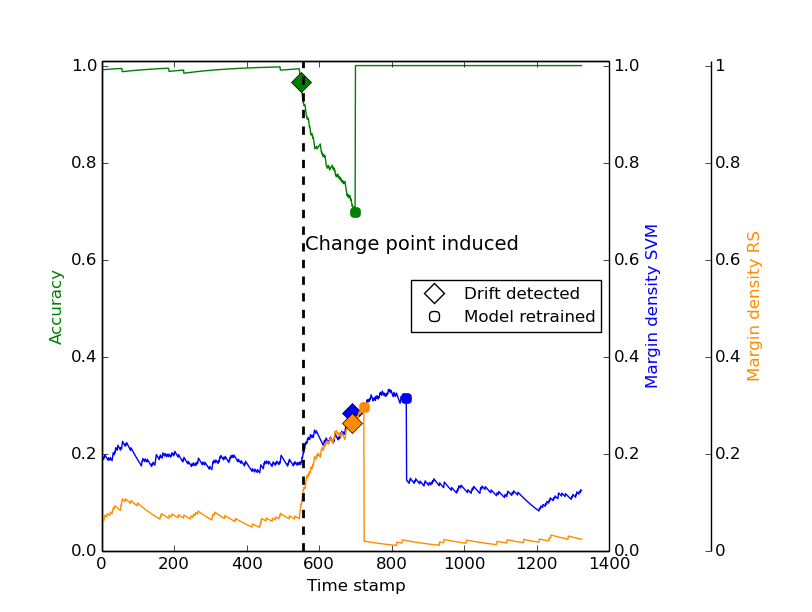}}
\subfloat[Musk]{\includegraphics[width=0.3\linewidth]{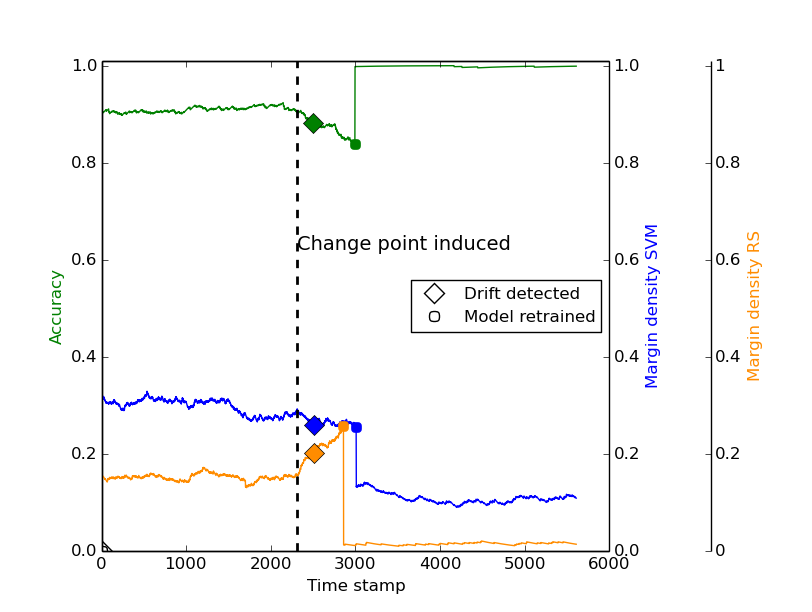}} \\
\subfloat[Wine]{\includegraphics[width=0.3\linewidth]{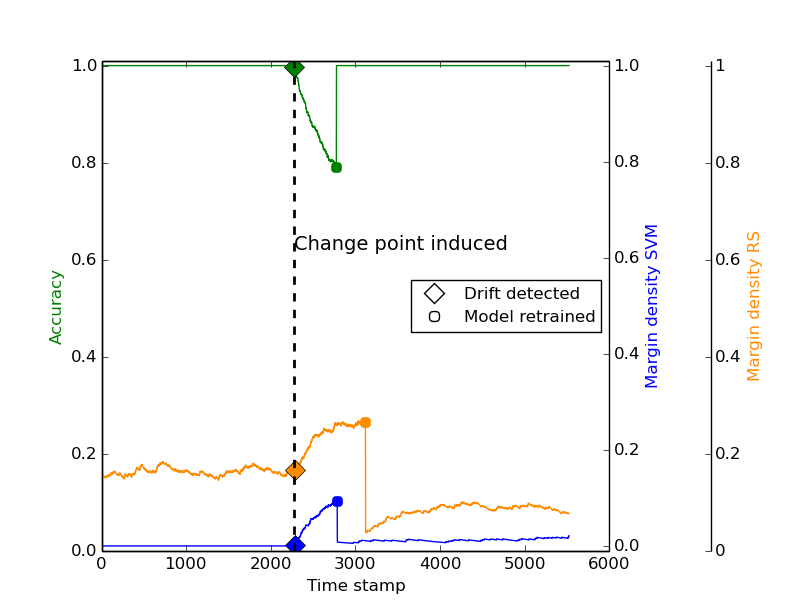}}
\subfloat[Bank]{\includegraphics[width=0.3\linewidth]{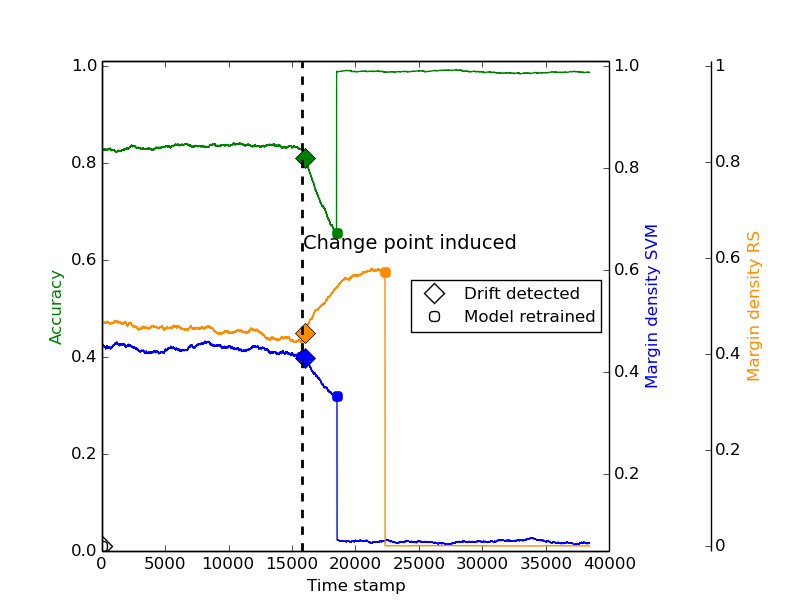}}
\subfloat[Adult]{\includegraphics[width=0.3\linewidth]{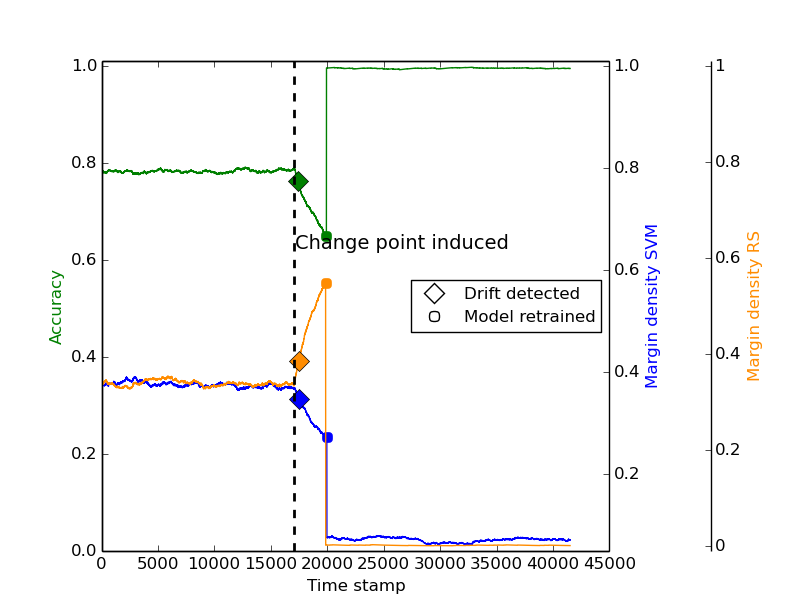}}
\caption{Accuracy (green), Margin Density for SVM (blue) and Margin Density for RS(orange), over time, in the \textit{Detectability} experiments. Drifts detected are denoted by diamonds and circles denote the retraining point.}
\label{fig:margindd}
\end{figure*}

The NoChange, AccTr, MD3-SVM, MD3-RS and the HDDDM methodologies, were analyzed in the \textit{Detectability} and the \textit{False alarm} experiments. The number of drifts detected, the false alarms raised and the accuracy over the stream, is reported in Table~\ref{tbl:induced_results}. A drift is detected if there is a significant change in the metric being tracked. This results in requesting of \textit{N} (Table~\ref{tbl:induced_data}) labeled samples, to confirm if the deviation leads to a drop in the accuracy. A false alarm is the result of a change signal, which after obtaining labeled samples was found to have no significant effect on the classification performance. In case a false alarm is reported, no retraining of the classifier takes place and the labeled samples are discarded. In the \textit{Detectability} experiments, there is exactly one true drift induced, which causes accuracy to drop after the midpoint of the stream. In case of the \textit{False alarm} experiments, the induced change does not affect performance and as such exactly one non relevant change is introduced in these experiments. 

The \textit{Detectability} experiment causes the model's performance to drop over time, which is evident from the low accuracy of the NoChange model, as seen in Table~\ref{tbl:induced_results}. This indicates the need for a drift detection methodology to deal with the induced drift. The AccTr approach directly monitors the classification performance with labeled samples, as such it serves as the gold standard for detecting drifts in our experiments. This approach detect exactly 1 drift in all 6 cases, indicating its robustness against false alarms. The AccTr, HDDDM and the MD3-SVM and MD3-RS techniques are all able to detect atleast one drift in these experiments (\textit{Detectability}) and as such are able to reach similar final accuracies. The MD3-RS and MD3-SVM are unsupervised methods and are still able to reach accuracy similar to the fully labeled AccTr approach (average difference of $<$1\% for both cases). This indicates the ability of the MD3 approaches to be used instead of the labeled approaches, without significantly compromising the prediction performance.

The resistance to false alarms is shown by the number of drifts detected in the \textit{False alarm}, in Table~\ref{tbl:induced_results}. Changes in these experiments do not cause a significant performance degradation. As such, a drift detected does not lead to retraining of the classifier, resulting in a false alarm. While the AccTr and both the MD3 approaches are resistant to such changes, the HDDDM approach signals it as a relevant change which needs further inspection. The HDDDM does not differentiate between the change in the top 25\% features vs the change in the bottom 25\% features, as it is a classifier agnostic technique which relies solely on tracking the changes in the raw feature values distribution. The HDDDM approach causes a higher false alarm rate, than the MD3 approaches, on both experiments. Another observation is that the MD3-SVM and MD3-RS show similar behavior, on average, with only a deviation of 0.25\% in accuracy and the exact same number of drifts detected. This shows the generic applicability of the margin density signal irrespective of the implementation technique used to compute it. 

Accuracy over time for the \textit{Detectability} experiments is shown in Figure~\ref{fig:dd}. After the \textit{ChangePoint}, a significant drop in accuracy is seen in all cases. The drift detection approaches are swift in recognizing this change and after retraining (seen as point after which accuracy starts to rise again) they are able to again provide high prediction performance. The NoChange approach (gray) does not detect any drift and as such the performance continues to degrade in this case. The MD3 approaches have accuracy trajectories close to the AccTr detector, indicating its use as a surrogate to the fully labeled approach. This accuracy is higher, in most cases, than the HDDDM approach. While the MD3 approach led to a false alarm on the Musk dataset only, the HDDDM approach is seen to trigger changes multiple times over the course of the stream. These changes are frequent (atleast 1, on average), and often occur without any correlation with the change in accuracy values. The MD3 approaches detect drifts in close proximity to the AccTr approach, seen by minimal lag between the green diamonds and, the orange and blue diamonds. The MD3-RS approach is more robust to change, which causes it to detect drifts with a delay in the Figure~\ref{fig:dd} d) and e). However, the drift detection for all approaches is close to the \textit{ChangePoint}, illustrating their ability to detect the true drift effectively. 

Progression of the Accuracy and Margin Density metrics over time, for the \textit{Detectability} experiment, is shown in Figure~\ref{fig:margindd}. A drop in the accuracy (green) is seen after the \textit{ChangePoint}. This is accompanied by significant spike in the MD3-RS (orange) metric. The metric MD3-SVM (blue) shows either a significant spike (in Figure~\ref{fig:margindd} a), b) and d)), or a drop in margin density (in Figure~\ref{fig:margindd} c), e) and f)), thereby justifying the need to compute the absolute deviation in MD as a signal for change. Also, the margin density metric, similar to the accuracy metric,  shows a high signal-to-noise ratio. It stays stable before the \textit{ChangePoint} and after the retraining is performed (indicated by circles in the plots), with a significant deviation only when drift occurs. This confirms that the drift detection is a result of the informativeness of the margin density metric, in detecting drifts, and not due to random variations in the data stream.

\subsection{Experiments on real world cybersecurity datasets exhibiting concept drift}
\label{sec:cdd}
Machine learning models deployed in real world applications operate in a dynamic environment where concept drift can occur at any time. Such drifts are not only plausible but in fact expected and rampant in the domain of cybersecurity, where attackers are constantly trying to generate data that degrades the classifier. In this section, 4 real world concept drift datasets are chosen from the domain of cybersecurity, as presented in Table~\ref{tbl:cdd_data}. These datasets are high dimensional and popularly used in machine learning literature, to test online binary classification models in a concept drifting environment. The spam and spamassassin datasets taken from \citep{katakis2010tracking, katakis2009adaptive}, represent the task of separating malicious spam email from legitimate ones. Phishing \citep{Lichman:2013} contains data about malicious web pages and the nsl-kdd dataset \citep{tavallaee2009detailed} is derived from the task of intrusion detection systems, which filters malicious network traffic. All the datasets were preprocessed by converting feature to numeric/binary types only, and by normalizing each feature value to the range of [0,1]. The final data characteristics are shown in Table~\ref{tbl:cdd_data}. These datasets exhibit concept drift, but the exact nature and location of the drifts is not known in advance. 

\begin{table}[t]
\centering
\caption{Description of real world concept drift datasets from cyber-security domain.}
\label{tbl:cdd_data}

\resizebox{0.45\textwidth}{!}{\begin{tabular}{|l|l|l|l|}
\hline
Dataset & \#Instances & \#Attributes & Chunk Size - N \\ \hline
spam & 6213 & 499 & 500 \\ \hline
spamassassin & 9324 & 499 & 500 \\ \hline
phishing & 11055 & 46 & 500 \\ \hline
nsl-kdd & 37041 & 122 & 2500 \\ \hline
\end{tabular}}
\end{table}

\begin{table*}[t]
\centering
\caption{Results on real world concept drift datasets.}
\label{tbl:cdd_results}
\begin{tabular}{|l|l|c|c|c|c|}
\hline
Dataset & Methodology & Accuracy & Drifts signaled & False alarms & Labeling \% \\ \hline
\multirow{5}{*}{spam} & NoChange & 57.5 & 0 & 0 & 0 \\ \cline{2-6} 
 & AccTr & 90.6 & 1 & 0 & 100 \\ \cline{2-6} 
 & MD3-SVM & 87.3 & 2 & 1 & 18.9 \\ \cline{2-6} 
 & MD3-RS & 89.2 & 2 & 1 & 18.9 \\ \cline{2-6} 
 & HDDDM & 80.7 & 2 & 1 & 18.9 \\ \hline
\multicolumn{1}{|c|}{\multirow{5}{*}{spamassassin}} & NoChange & 67.0 & 0 & 0 & 0 \\ \cline{2-6} 
\multicolumn{1}{|c|}{} & AccTr & 92.8 & 1 & 0 & 100 \\ \cline{2-6} 
\multicolumn{1}{|c|}{} & MD3-SVM & 92.8 & 2 & 1 & 12.6 \\ \cline{2-6} 
\multicolumn{1}{|c|}{} & MD3-RS & 92.8 & 2 & 1 & 12.6 \\ \cline{2-6} 
\multicolumn{1}{|c|}{} & HDDDM & 92.6 & 4 & 3 & 21.9 \\ \hline
\multirow{5}{*}{phishing} & NoChange & 86.9 & 0 & 0 & 0 \\ \cline{2-6} 
 & AccTr & 92.9 & 2 & 0 & 100 \\ \cline{2-6} 
 & MD3-SVM & 91.7 & 1 & 0 & 5.3 \\ \cline{2-6} 
 & MD3-RS & 90.8 & 1 & 0 & 5.3 \\ \cline{2-6} 
 & HDDDM & 92.8 & 4 & 2 & 21.3 \\ \hline
\multirow{5}{*}{nsl-kdd} & NoChange & 79.6 & 0 & 0 & 0 \\ \cline{2-6} 
 & AccTr & 90.1 & 1 & 0 & 100 \\ \cline{2-6} 
 & MD3-SVM & 89.4 & 1 & 0 & 7.9 \\ \cline{2-6} 
 & MD3-RS & 89.9 & 1 & 0 & 7.9 \\ \cline{2-6} 
 & HDDDM & 87.2 & 2 & 1 & 15.8 \\ \hline
\end{tabular}
\end{table*}

\begin{figure*}[t]
\centering
\subfloat[spam]{\includegraphics[width=0.4\linewidth]{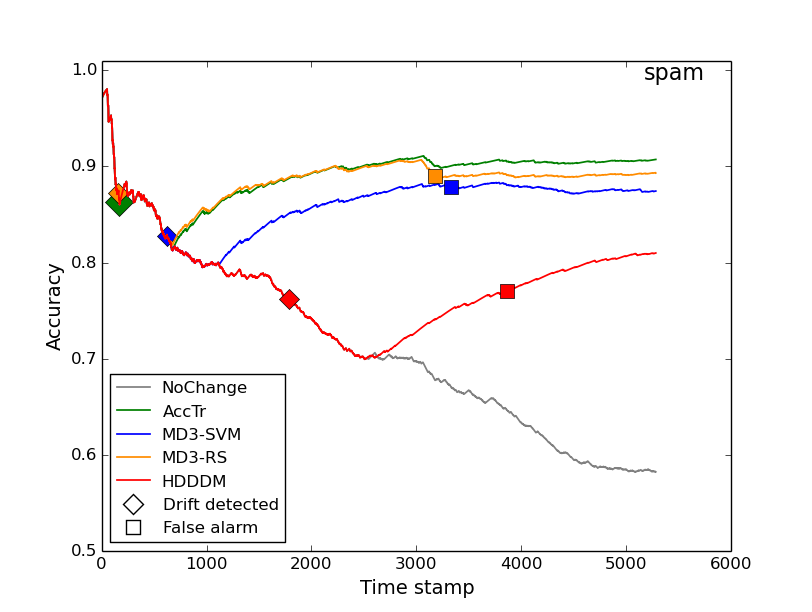}}
\subfloat[spamassassin]{\includegraphics[width=0.4\linewidth]{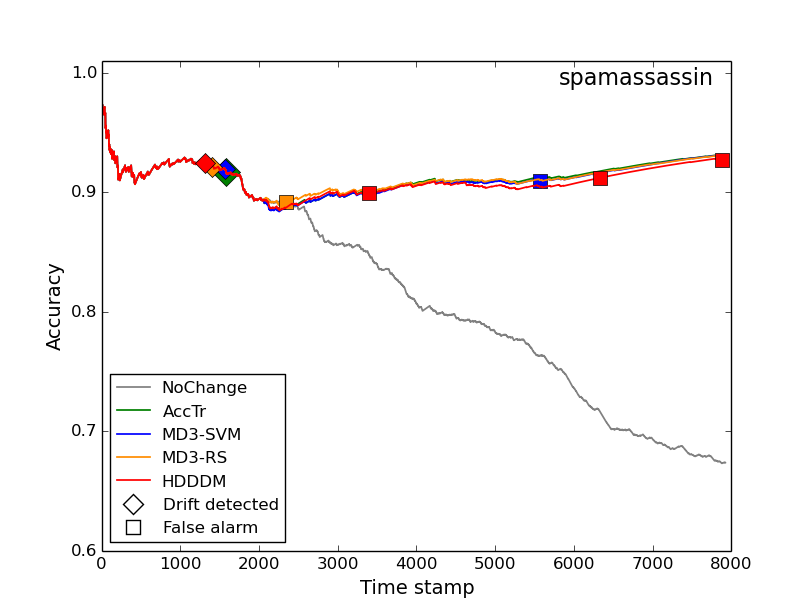}} \\
\subfloat[phishing]{\includegraphics[width=0.4\linewidth]{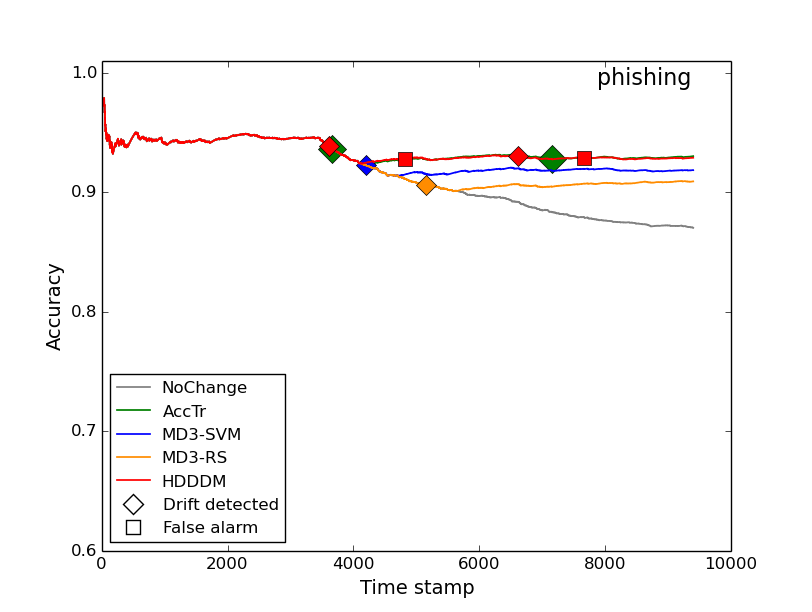}}
\subfloat[nsl]{\includegraphics[width=0.4\linewidth]{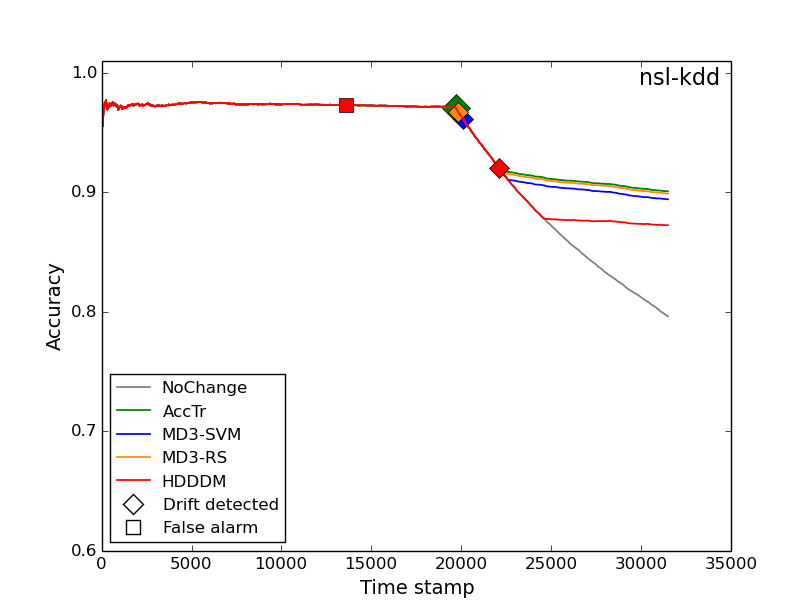}}
\caption{Accuracy over time for the NoChange(gray), AccTr(green), MD3-SVM(blue), MD3-RS(orange) and the HDDDM(red) approach on real world concept drift datasets. True drifts detected are shown as diamonds and squares represent false alarms.}
\label{fig:cdd_real}
\end{figure*}

The NoChange, AccTr, MD3-SVM, MD3-RS and HDDDM methodologies are evaluated on the 4 datasets. The metrics used for evaluation are: Accuracy of the stream, number of drifts signaled, number of false alarms and the total percentage of samples which were requested to be labeled. Accuracy determines the predictive performance of the online system. Number of drifts signaled indicates sensitivity to change. False alarms occur when a drift is signaled, but upon obtaining labeled samples it was found that performance has not significantly degraded, thus requiring no retraining. Since the location and number of drifts is not known in advance, the false alarm refers to cases only which did not lead to a retraining of the models, causing the requested labels to go wasted. False alarms are 0 for the case of the AccTr approach as this approach directly tracks drop in accuracy, unlike the other unlabeled techniques. False alarms refers to situation where Equation (\ref{eqn:changemd}) is triggered due a suspected drift, but upon receiving labeled samples we confirm that retraining is not needed (Equation (\ref{eqn:changeacc})), as the accuracy has not degraded significantly. The labeling\% indicates the cost expended by the methodology (in term of labels requested) and is directly related to the number of false alarms, as every alarm leads to requesting of Chunk Size- \textit{N} (shown in Table~\ref{tbl:cdd_data}) samples to be labeled. A high accuracy, high drift detection, low false alarm and low labeling\% is desirable. 

In all cases, the accuracy of the NoChange approach is significantly lower than the other drift detection techniques, as observed from Table~\ref{tbl:cdd_results}. This confirms the drifting nature of the datasets and the need for drift detection. The accuracy obtained by the margin density methodologies is close to the fully labeled AccTr approach, with MD3-SVM having an average deviation of 1.3\% and the MD3-RS having a 0.9\% deviation, only. This indicates the ability of these techniques to detect drift as good as a labeled drift detection mechanism. The labeling requirement for the MD3 approaches is 88.3\% less than the AccTr approach, which relies on a totally labeled stream for performing its computation. 

The HDDDM approach, on average, performs poorly compared to the MD3 approaches, and also needs 8.3\% more labeling. This is a result of its high false alarm rate. False alarms are harmful as it causes the system to react to every change in the data distribution, noise or otherwise, making adaptiveness a hassle, rather than a solution. Especially in the domain of streaming cybersecurity applications, an overly responsive system is a serious problem, as it is vulnerable to malicious manipulation of the training process. Also, labeling is a time consuming and expensive task. A system which frequently requires manual intervention is less likely to be trusted and can cause experts to disregard its warnings. The HDDDM approach needed 1250 more labeled samples than the MD3 approaches, on average, due to its increased false alarm rate. The MD3 approach on the other hand, signals change only when it would affect the system performance directly, making it suitable as a self-guided automatic monitoring system for drift detection and malicious activity on the models.

Progression of accuracy over the 4 datasets, is shown in Figure~\ref{fig:cdd_real}. It is seen that the MD3 methodologies and the AccTr approach converge and behave similarly over time. This indicates that MD3 could be used as a replacement for a fully labeled drift detection approach. The similarity in the MD3-SVM (blue) and the MD3-RS(orange) approaches, in terms of the accuracy progression and the number of drifts/false alarms detected, indicates the general applicability of the margin density metric as a drift indicator metric. The margin density metric provides good performance irrespective of the type of machine learning technique it is implemented on, making it classification algorithm independent. 

The HDDDM (red) approach performs worse than the other methods on the spam and the nsl-kdd datasets. This is attributed to the delay in its drift detection. In all datasets, the number of false alarms (squares) is high for the HDDDM approach, which translates to increased labeling expenditure as shown in Table~\ref{tbl:cdd_results}. The MD3 approaches signal 1 false alarm in the spam and spamassassin datasets. In the spam dataset, the MD3-SVM and the MD3-RS approaches signal change at timestamps 3829 and 3676 respectively (Figure~\ref{fig:cdd_real}a), which are reported as false alarms. However, these are not without basis, there is a small drop in the accuracy at these points, albeit not enough to warrant concept drift recovery. The margin density metric is sensitive to changes in accuracy, but is still robust compared to other feature tracking approaches (HDDDM), which seem to signal changes without any correlation to accuracy degradation. Machine learning based cybersecurity systems will benefit from MD3, due to its increased reliability for change detection, robustness to irrelevant changes, and the reduced need for manual intervention.

\subsection{Effects of varying the detection model and the margin width ($\theta_{margin}$)}
\label{sec:modelsmargins}

In all the experiments so far, the detection model was taken to be a Linear SVM with regularization parameter of C=1, for the MD3-SVM technique, and a random subspace ensemble with C4.5 Decision Trees as its base models, for the MD3-RS technique. Moreover, the margin width ($\theta_{margin}$), was kept fixed at 0.5, based on intuition. The effects of varying these settings on the detection capabilities of the MD3 framework, are presented in this section. 

\subsubsection{Effects of varying the detection model}
\label{sec:models} 

\begin{figure*}[t]
\centering
\subfloat[spam]{\includegraphics[width=0.4\linewidth]{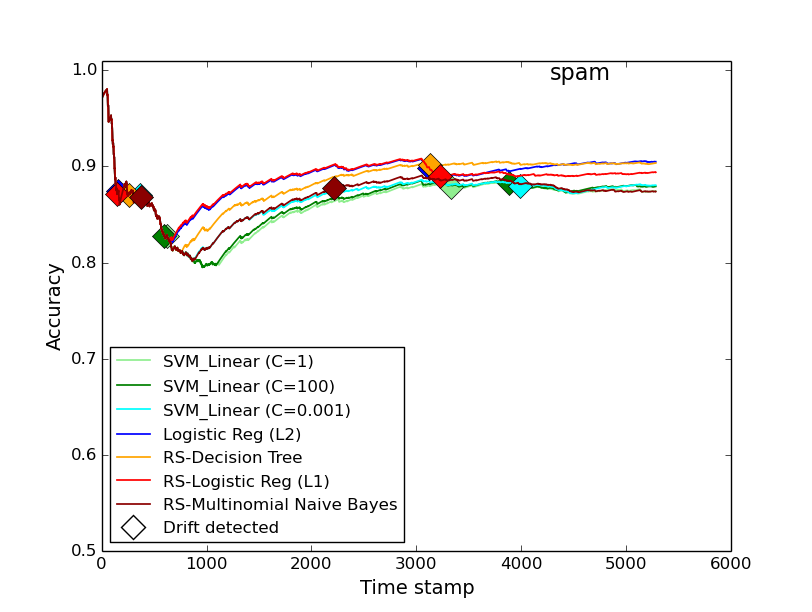}}
\subfloat[spamassassin]{\includegraphics[width=0.4\linewidth]{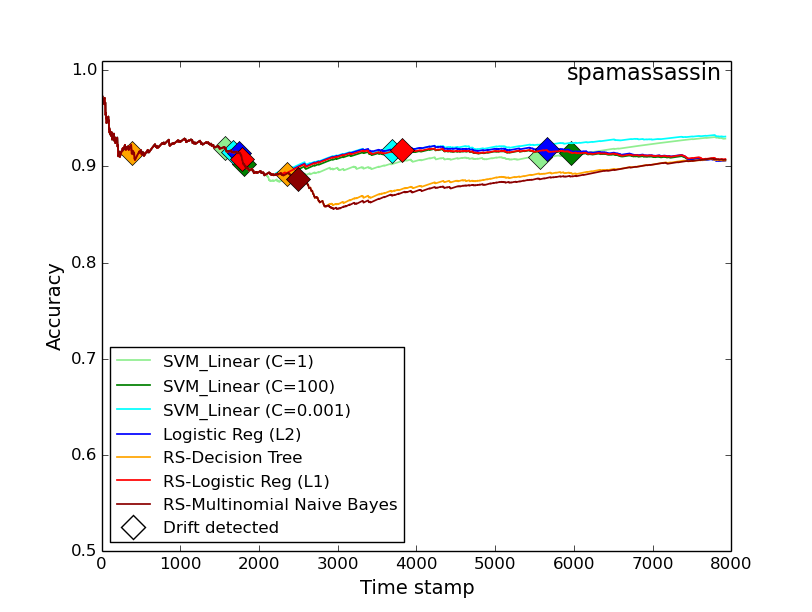}} \\
\subfloat[phishing]{\includegraphics[width=0.4\linewidth]{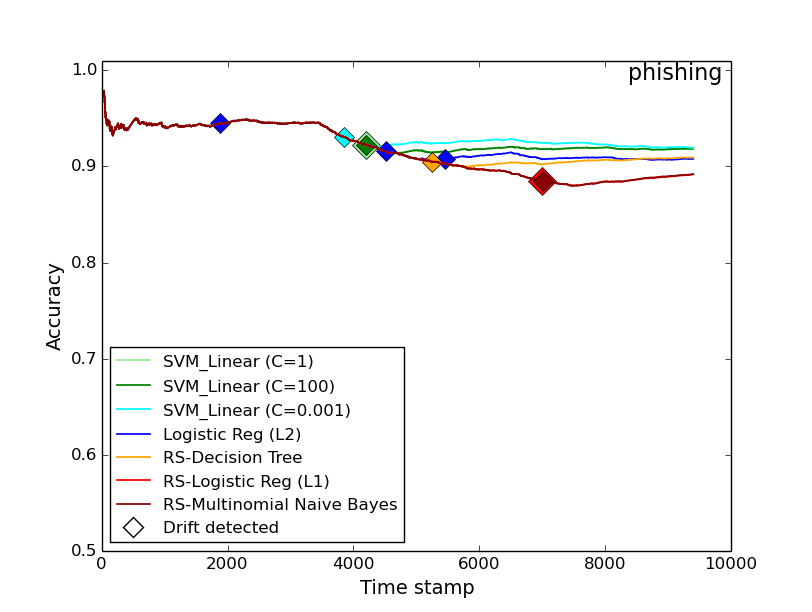}}
\subfloat[nsl]{\includegraphics[width=0.4\linewidth]{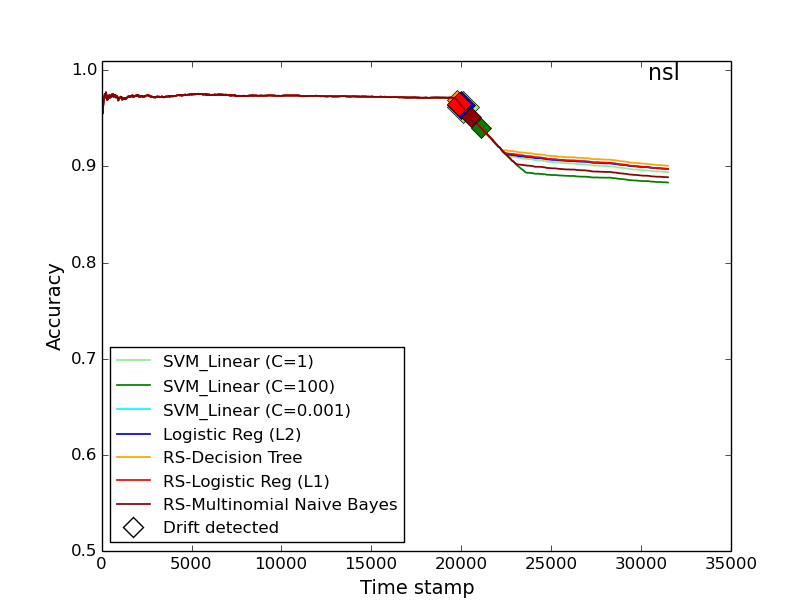}}
\caption{Effect of varying the detection model on the drift detection and the prediction accuracy, over time. }
\label{fig:models}
\end{figure*}

The 4 real world datasets from Section~\ref{sec:cdd} are evaluated here using the following 5 additional detection models: linear SVM with C=100 (high regularization constant), linear SVM with C=0.001(Low regularization constant), logistic regression with L2-penalty, random subspace ensemble with logistic regression (L1-penalty) models and random subspace ensemble with multinomial naive Bayes models. The results of these experiments are presented in Figure~\ref{fig:models}. All models were evaluated using the scikit-learn machine learning library \citep{scikit-learn}. 

Experimental results of varying the detection models are presented in Figure~\ref{fig:models}. It is observed that varying the underlying model has no significant effect on the detection capabilities of the MD3 methodology. A Friedman's non parametric test \citep{demvsar2006statistical} on the final accuracy values for all the 4 datasets, showed that there is no statistically significant difference between the performance of the different models, at a p-value of 0.05. The number of drifts detected and the relative position of the detection are also observed to be close in a majority of the cases. 

These results are inline with our intuition in Section~\ref{sec:motivation}, where we postulate that the margin density signal could be a general,  model independent indicator of change that could be applied to any robust classifier model, which distributes classification importance weights among its features. The models in the experiments described here are all robust classifiers, which distribute feature weights, and as such perform similarly under the MD3 framework. The results of applying the MD3 methodology on a non-robust classifier is shown in Table~\ref{tbl:l1}, where a logistic regression model with L1-penalty is used. The margin for a logistic regression model is defined as described in Section~\ref{sec:computing_rs}, with the probability obtained from the posterior estimates of the classifier. This classifier tends to minimize the number of features used in the final models, making it unsuitable for the MD3 methodology. The results of Table~\ref{tbl:l1} show that the model does not detect any drifts for the phishing dataset and also performs significantly worse on the spam and the spamassassin dataset. This is because the L1-penalty model tends to minimize the number of features used, which violates the central premise of coupled features detection (Section~\ref{sec:motivation}) that the MD3 model relies on. However, it can be seen from Figure~\ref{fig:models} that the same L1-penalty based logistic regression model when used with the random subspace ensemble can be effective for usage under the MD3 framework. The MD3 methodology, with its ability to use models of explicit and non-explicit margins, can therefore be applied as a general detection scheme irrespective of the base models used. 

\begin{table}[]
\centering
\caption{Results of using logistic regression (L1-penalty) as the detection model for MD3.}
\label{tbl:l1}
\begin{tabular}{|l|c|c|}
\hline
Dataset & \multicolumn{1}{l|}{Accuracy} & \multicolumn{1}{l|}{Drifts signaled} \\ \hline
spam & 87.8 & 2 \\ \hline
spamassassin & 85.3 & 2 \\ \hline
phishing & 86.9 & 0 \\ \hline
nsl-kdd & 89.3 & 1 \\ \hline
\end{tabular}
\end{table}

\subsubsection{Effects of varying the margin width ($\theta_{margin}$)}
\label{sec:marginwidths}

\begin{figure*}[t]
\centering
\subfloat[spam]{\includegraphics[width=0.38\linewidth]{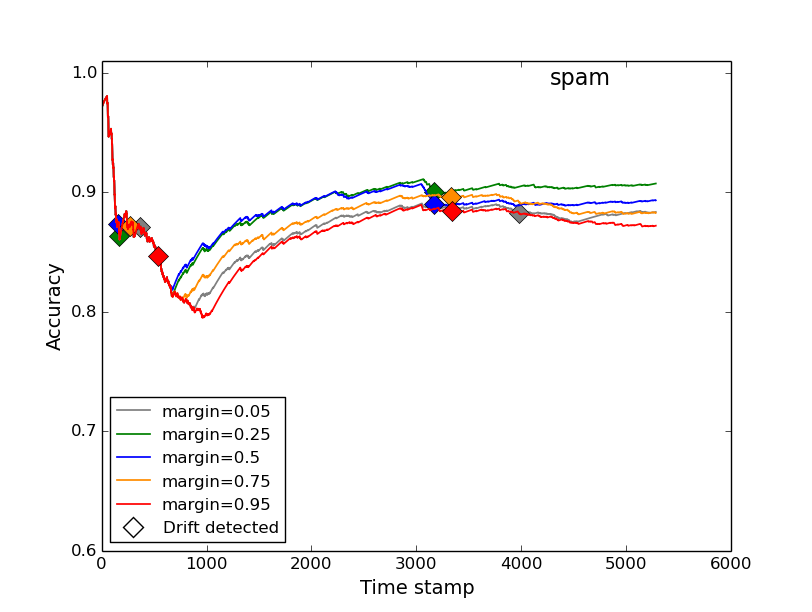}}
\subfloat[spamassassin]{\includegraphics[width=0.38\linewidth]{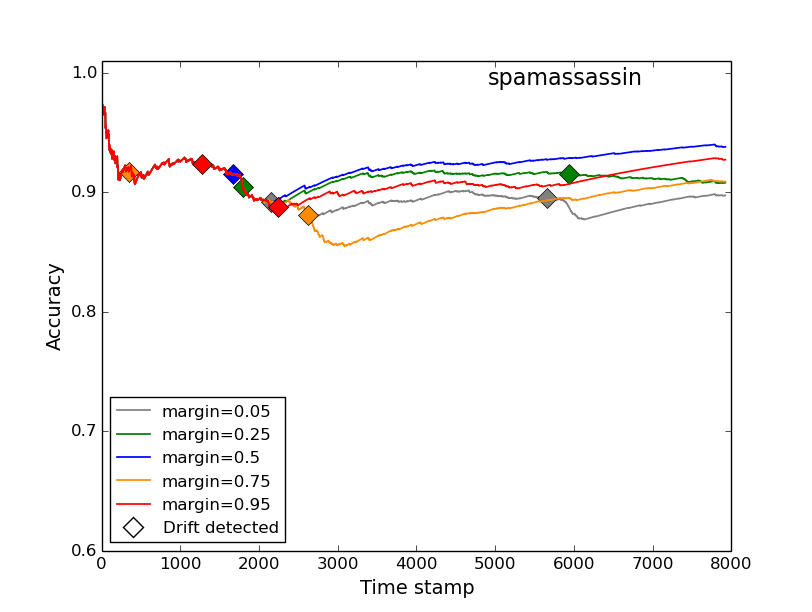}} \\
\subfloat[phishing]{\includegraphics[width=0.38\linewidth]{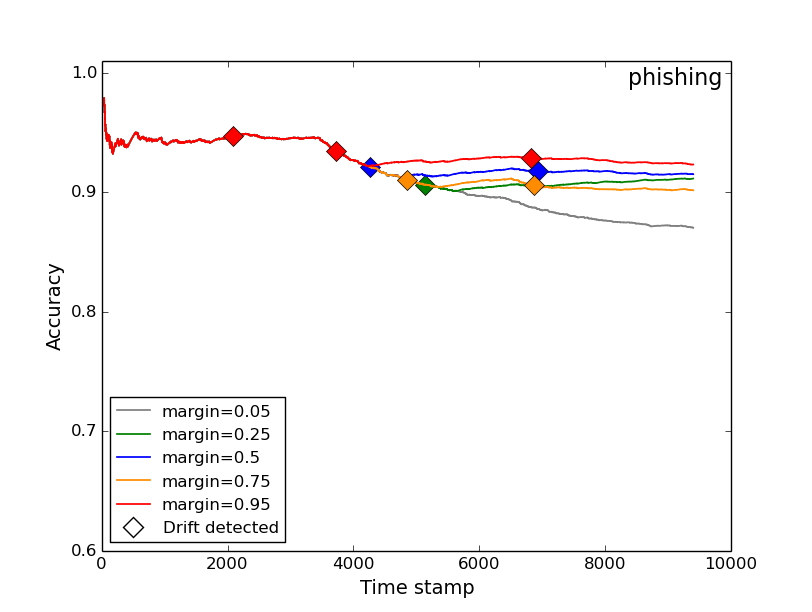}}
\subfloat[nsl]{\includegraphics[width=0.38\linewidth]{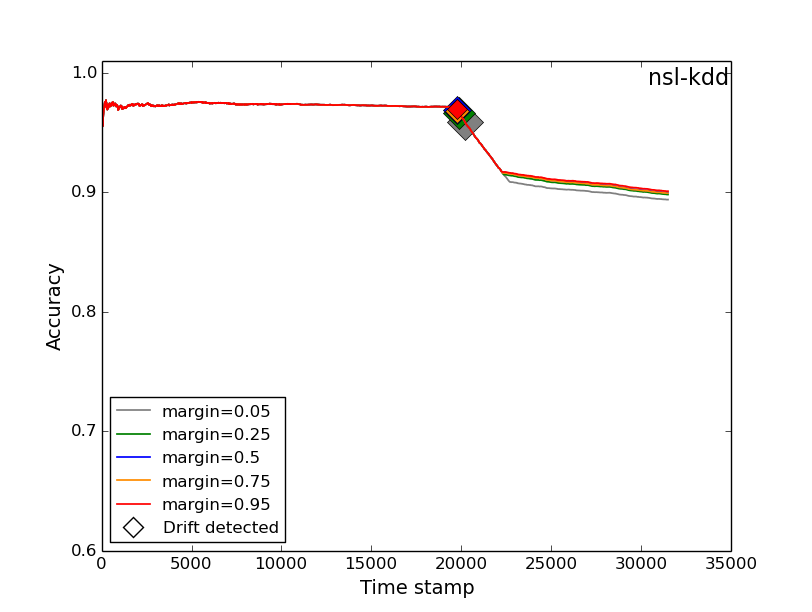}}
\caption{Effects of varying the margin width ($\theta_{margin}$) on the drift detection process of MD3.}
\label{fig:marginw}
\end{figure*}

\begin{figure*}[t]
\centering
\subfloat[phishing]{\includegraphics[width=0.4\linewidth]{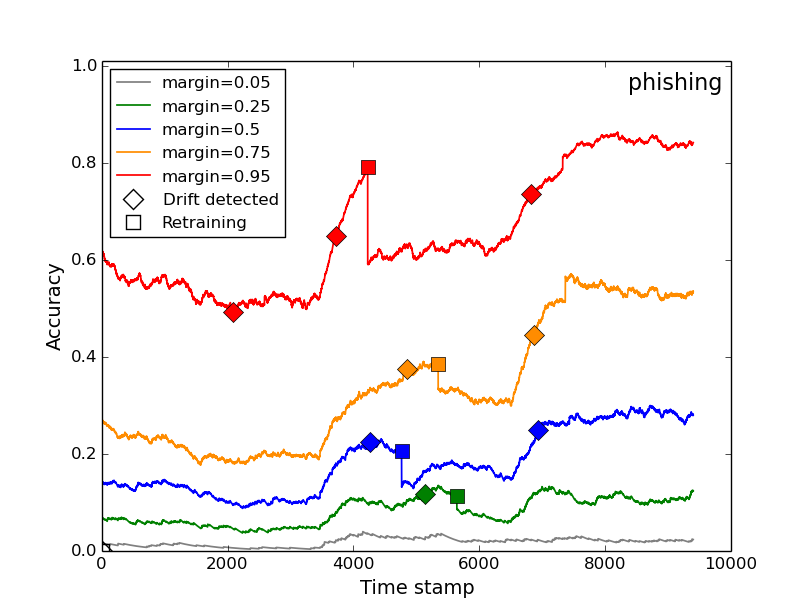}}
\subfloat[nsl]{\includegraphics[width=0.4\linewidth]{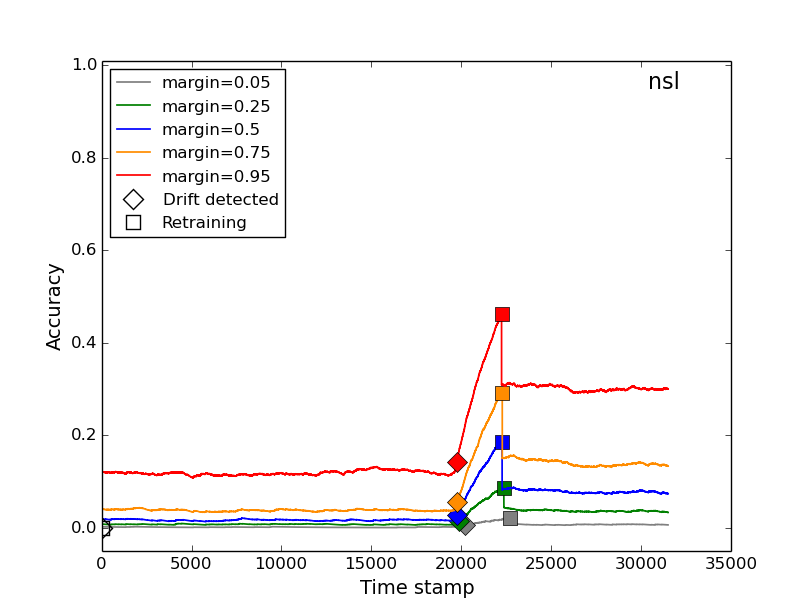}}
\caption{Margin density metric over time, for different values of ($\theta_{margin}$).}
\label{fig:marginplot}
\end{figure*}

In case of the MD3-RS approach, the concept of margin is defined by the margin width parameter $\theta_{margin}$, which was taken as 0.5 in all the experiments thus far. Experiments in the previous section demonstrate that varying the underlying detection model has no significant effect on the detection capabilities of the MD3 framework. In this section, we evaluate the effect of varying the parameter $\theta_{margin}$ for a MD3-RS model, which uses C4.5 decision tree models as its base classifier type.

The results of the varying the $\theta_{margin}$ are shown in Figure~\ref{fig:marginw}. It is seen that the choice of the parameter $\theta_{margin}$ does not have a significant effect on the final results, as all accuracy plots follow a similar trajectory in time. The only failure case is seen in case of a $\theta_{margin}$=0.05 for the \textit{phishing} dataset (Figure~\ref{fig:marginw} c) gray). At this margin width, the samples captured are insufficient to detect drift effectively. The margin density signal is depicted in Figure~\ref{fig:marginplot}, for the phishing and the nsl-kdd dataset. It is seen that, for the phishing dataset at a $\theta_{margin}$=0.05 the signal fails to detect a drift. For all other margin values, although the absolute signal magnitude is different, they are all effective in detecting change and do so at nearly the same location. This can be attributed to the reference distribution learning component of the MD3 algorithm, which learns the expected margin density via cross validation on the training dataset. Subsequent changes tracked are relative to the reference distribution, making them effective even when margin width changes. To maintain robustness to change and to ensure that drift detection is effective, a $\theta_{margin}$ in the range of 0.25-0.75 is suggested. Further tuning can be done based on the desired level of sensitivity required for the application.

\subsection{Experimental results on benchmark concept drift datasets}
\label{sec:benchmark}

\begin{figure*}[t]
\centering
\subfloat[Electricity Market (EM) dataset]{\includegraphics[width=0.4\linewidth]{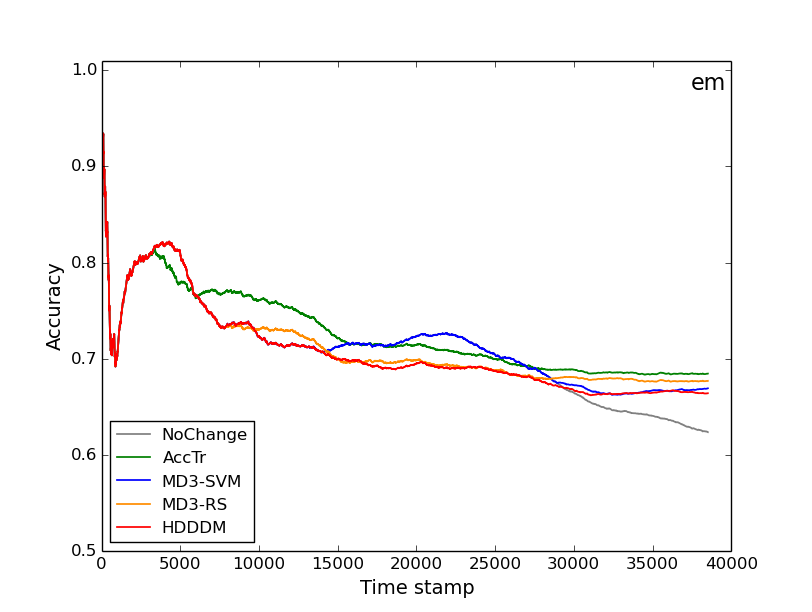}}
\subfloat[Covertype dataset]{\includegraphics[width=0.4\linewidth]{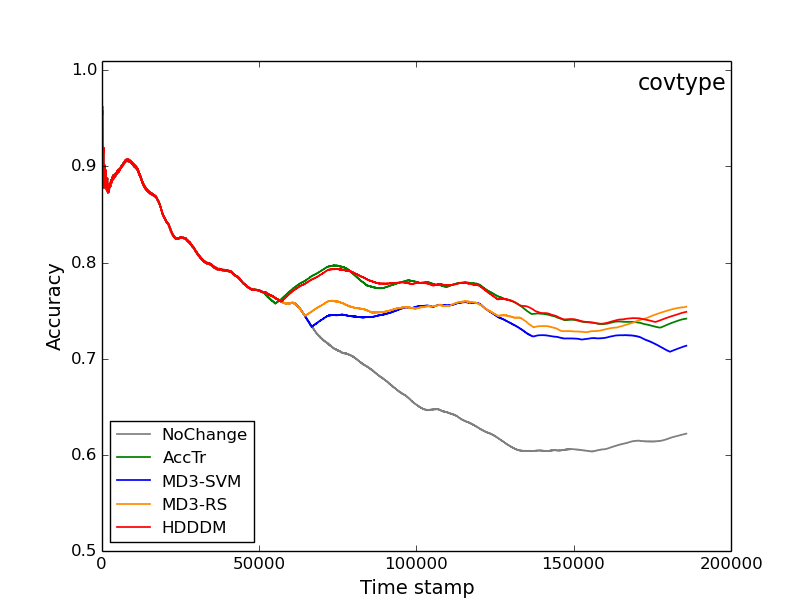}}
\caption{Accuracy over time for the NoChange(gray), AccTr(green), MD3-SVM(blue), MD3-RS(orange) and the HDDDM(red) approach on real world benchmark concept drift datasets.}
\label{fig:benchmark}
\end{figure*}

\begin{table*}[]
\centering
\caption{Results on benchmark real world concept drift datasets.}
\label{tbl:benchmark}
\begin{tabular}{|c|l|c|c|c|}
\hline
\multicolumn{1}{|l|}{Dataset} & Methodology & \multicolumn{1}{l|}{Accuracy} & \multicolumn{1}{l|}{Drifts signaled} & \multicolumn{1}{l|}{Labeling \%} \\ \hline
\multirow{5}{*}{\begin{tabular}[c]{@{}c@{}}EM\\ (\#Instances = 45312, \\ \#Features = 8)\end{tabular}} & NoChange & 62.3 & 0 & 0 \\ \cline{2-5} 
 & AccTr & 68.4 & 2 & 100 \\ \cline{2-5} 
 & MD3-SVM & 66.9 & 2 & 13 \\ \cline{2-5} 
 & MD3-RS & 67.7 & 2 & 13 \\ \cline{2-5} 
 & HDDDM & 66.4 & 4 & 26 \\ \hline
\multirow{5}{*}{\begin{tabular}[c]{@{}c@{}}Covtype \\ (\#Instances = 218515, \\ \#Features = 54)\end{tabular}} & NoChange & 62.2 & 0 & 0 \\ \cline{2-5} 
 & AccTr & 74.2 & 16 & 100 \\ \cline{2-5} 
 & MD3-SVM & 71.3 & 18 & 24.2 \\ \cline{2-5} 
 & MD3-RS & 75.4 & 22 & 29.6 \\ \cline{2-5} 
 & HDDDM & 74.9 & 25 & 33.6 \\ \hline
\end{tabular}
\end{table*}

The MD3 methodology is evaluated on two popular real world data streams here - the Electricity market (EM) dataset and the Covertype dataset (Covtype).  These are widely used datasets for benchmark test of concept drift handling systems \citep{gama2004learning, goncalves2014comparative}. The electricity market dataset represents pricing data collected from New South Wales, Australia, which fluctuates based on the supply and the demand components of the market \cite{gama2004learning, goncalves2014comparative}. The Covertype dataset consists of forest cover data and is a multi-class dataset \cite{bifet2010fast}. This dataset was reduced to a binary class prediction problem by considering only the class labels of 1 and 2. Both datasets were pre-processed by converting features to numeric values, normalized in the range of [0,1]. Chunk size \textit{N} of 2500 was chosen for evaluation of both datasets. These datasets have unknown type and location of the drift, and as such serve as real world benchmarks for evaluating concept drift techniques. 

The results of Table~\ref{tbl:benchmark} show that although both datasets have unknown concept drift, they benefit from drift handling, as the accuracy of the \textit{NoChange} approach is worse than that of the other techniques (Figure~\ref{fig:benchmark}). The MD3 approaches were found to have similar accuracy to the fully labeled \textit{AccTr} approach ($\Delta=1.1\%$ for EM and $\Delta=0.85\%$ for Covtype). The MD3 approach was also seen to signal fewer drifts than the HDDDM approach, leading to half the labeling budget in case of EM and \textit{$3/4^{th}$} in case of the Covtype dataset, for the same resulting accuracy, on average. 

\section{How the MD3 compares to other margin based drift detection techniques?}
\label{sec:marginanalysis}

\begin{table*}[t]
\centering
\caption{Characteristics of synthetic data generator used for comparing effects of the different margin based change detection metrics. (Table~\ref{tbl:marginanalysis}). }
\label{tbl:marginsynthetic}
\begin{tabular}{|l|}
\hline

\textit{\textbf{Before drift (1-500 Samples):}} \\ \\
$
X_{Class 1}\sim \mathcal{N}(\mu= \{0.5,...(5\ features),0.85,0.85,...,0.85\ (Upto\ 20 features)\},\sigma=0.1^2)
$
\\
$
X_{Class 2}\sim \mathcal{N}(\mu= \{0.5,...\ (5\ features),0.15,0.15,...,0.15\ (Upto\ 20 features)\},\sigma=0.1^2)
$ \\
\hline

\textit{\textbf{After Drift (501-1000 Samples):}} \\ \\
\textit{No Change:} \\
$
X_{Class 1}\sim \mathcal{N}(\mu= \{0.5,...(5\ features),0.85,0.85,...,0.85\ (Upto\ 20 features)\},\sigma=0.1^2)
$
\\

\textit{Change Upto feature i:} \\
$
X_{Class 2}\sim \mathcal{N}(\mu= \{0.75,...,0.75\ (i\ features), 0.15,...,0.15\ (Upto\ 20 features)\},\sigma=0.1^2)
$ \\
\hline
\end{tabular}
\end{table*}

\begin{table}[t]
\centering
\caption{Results of change detection metrics $\Delta$Err, $\Delta$MD, $\Delta$Uncertain and $\Delta$HD, on varying intensities of drift on synthetic data. Features 1-5 are irrelevant to the classification. Bold entries represent first indication of change for each of the metrics.}
\label{tbl:marginanalysis}
\begin{tabular}{lllll}
\hline
\multicolumn{1}{|l|}{\begin{tabular}[c]{@{}l@{}}\# of features\\   affected (\textit{i})\end{tabular}} & \multicolumn{1}{l|}{$\Delta$Err} & \multicolumn{1}{l|}{$\Delta$MD} & \multicolumn{1}{l|}{\begin{tabular}[c]{@{}l@{}}$\Delta$Uncertain\end{tabular}} & \multicolumn{1}{l|}{$\Delta$HD} \\ \hline
\multicolumn{1}{|l|}{\textit{1 (Irrelevant)}} & \multicolumn{1}{l|}{0} & \multicolumn{1}{l|}{0} & \multicolumn{1}{l|}{0} & \multicolumn{1}{l|}{\textbf{0.13}} \\ \hline
\multicolumn{1}{|l|}{\textit{3 (Irrelevant)}} & \multicolumn{1}{l|}{0} & \multicolumn{1}{l|}{0} & \multicolumn{1}{l|}{0} & \multicolumn{1}{l|}{0.17} \\ \hline
\multicolumn{1}{|l|}{\textit{5 (Irrelevant)}} & \multicolumn{1}{l|}{0} & \multicolumn{1}{l|}{0} & \multicolumn{1}{l|}{0} & \multicolumn{1}{l|}{0.22} \\ \hline
\multicolumn{1}{|l|}{6} & \multicolumn{1}{l|}{0} & \multicolumn{1}{l|}{0} & \multicolumn{1}{l|}{\textbf{0.05}} & \multicolumn{1}{l|}{0.23} \\ \hline
\multicolumn{1}{|l|}{7} & \multicolumn{1}{l|}{0} & \multicolumn{1}{l|}{0} & \multicolumn{1}{l|}{0.2} & \multicolumn{1}{l|}{0.26} \\ \hline
\multicolumn{1}{|l|}{8} & \multicolumn{1}{l|}{0} & \multicolumn{1}{l|}{0} & \multicolumn{1}{l|}{0.2} & \multicolumn{1}{l|}{0.28} \\ \hline
\multicolumn{1}{|l|}{9} & \multicolumn{1}{l|}{0} & \multicolumn{1}{l|}{0} & \multicolumn{1}{l|}{0.25} & \multicolumn{1}{l|}{0.3} \\ \hline
\multicolumn{1}{|l|}{10} & \multicolumn{1}{l|}{0} & \multicolumn{1}{l|}{0} & \multicolumn{1}{l|}{0.25} & \multicolumn{1}{l|}{0.33} \\ \hline
\multicolumn{1}{|l|}{11} & \multicolumn{1}{l|}{0} & \multicolumn{1}{l|}{\textbf{0.5}} & \multicolumn{1}{l|}{0.4} & \multicolumn{1}{l|}{0.35} \\ \hline
\multicolumn{1}{|l|}{12} & \multicolumn{1}{l|}{\textbf{0.5}} & \multicolumn{1}{l|}{0.5} & \multicolumn{1}{l|}{0.45} & \multicolumn{1}{l|}{0.37} \\ \hline
\multicolumn{1}{|l|}{13} & \multicolumn{1}{l|}{0.5} & \multicolumn{1}{l|}{0.5} & \multicolumn{1}{l|}{0.45} & \multicolumn{1}{l|}{0.39} \\ \hline
\multicolumn{1}{|l|}{14} & \multicolumn{1}{l|}{0.5} & \multicolumn{1}{l|}{0.5} & \multicolumn{1}{l|}{0.35} & \multicolumn{1}{l|}{0.41} \\ \hline
\multicolumn{1}{|l|}{15} & \multicolumn{1}{l|}{0.5} & \multicolumn{1}{l|}{0.5} & \multicolumn{1}{l|}{0.4} & \multicolumn{1}{l|}{0.43} \\ \hline
 &  &  &  &  \\
 &  &  &  & 
\end{tabular}
\end{table}

The ability of the margin density signal to effectively ignore changes to the unlabeled data, which do not affect the classification performance, makes its usage attractive as an unsupervised drift indicator. This is because, unlike the feature based change detectors like HDDDM, the margin density (MD) approach implicitly includes the model in the drift detection process. Other unlabeled drift detection techniques developed in literature \citep{dries2009adaptive,lindstrom2013drift,dredze2010we,zliobaite2010change}
 and described in Section~\ref{sec:lr_model}, also incorporate the notion of a margin. However, these techniques differ from the MD3 approach in the signal being tracked. The MD3 technique tracks the margin density (MD) signal, which is the expected number of samples in the uncertain regions of a classifier. The other margin based techniques of Section~\ref{sec:lr_model} essentially track the average change in the uncertainty of samples over time. The difference between the two paradigms is subtle, and the ability to specify a fixed margin, before deployment, is responsible for the efficacy of the margin density approach. To elucidate the difference between these paradigms and their implications, a synthetic experiment is designed in this section. 

In order to understand the effects of the different margin based techniques, an experiment similar to that in Section~\ref{sec:proofofconcept} is performed. A synthetic 20 dimensional dataset is generated. The dataset has 1000 samples, with concept drift occurring at the midpoint (500 samples). The characteristics of the data are shown in Table~\ref{tbl:marginsynthetic}. The dataset has 20 dimensions, the first 5 of which are made irrelevant to the classification task, by assigning them the same distributions for the two classes. After the midpoint, drift is induced in the dataset by changing Class 2's feature distribution. The feature distributions are incrementally changed upto feature \textit{i}, by changing their mean values. The remaining 15 features (feature 6-20) are all relevant to the classification task and the effect of changing them gradually is analyzed via this experiment. The effects of these changes are evaluated by training a model on the first 500 samples and evaluating change metrics on the remaining 500 samples after the drift. We chose the random subspace ensemble with decision trees as its base models (MD3-RS), for the experiments here. 

The effects of changing the feature values of the synthetic 20-Dimensional dataset, is presented in Table~\ref{tbl:marginanalysis}. The change in the training and testing error ($\Delta$Err), the margin density metric ($\Delta$MD), the average uncertainty  ($\Delta$Uncertain) and the Hellinger Distance ($\Delta$HD), are compared. Changing feature 1-5 have no significant effect on the classification error, as these features are irrelevant from a classifier's perspective. The HD metric, owing to its model agnostic measurements, is unable to see this distinction and as such a change is seen for the $\Delta$HD metric in Table~\ref{tbl:marginanalysis}. Both the margin density technique and the uncertainty tracking techniques, are robust against such changes. The advantage of the MD technique, over the traditional uncertainty tracking techniques, is seen in case of the changes in the relevant features (6-20). A robust classifier, such as the random subspace ensemble, is inherently capable of providing high predictive performance, even if a few of the relevant features drift. Unless a majority of the features change at the same time, the robust classifier is unaffected by changes to a few features. The traditional uncertainty tracking techniques fail to distinguish between critical changes and changes which can be inherently managed by a robust classifier. By tracking the posterior probability estimates of the classifier models, the uncertainty based techniques could effectively ignore changes to irrelevant data (similar to the MD3 approach), but they still lead to additional false alarms when compared to the margin density approach. The MD approach limits the critical area of uncertainty being monitored and as such it limits the tracking to the most critical drifts only. This is seen in Table~\ref{tbl:marginanalysis}, where the MD metric is affected after 11 features drift simultaneously, as opposed to 6 features for the \textit{Uncertainty} tracking techniques. The actual fully labeled technique would signal drift after 12 features are affected. The MD signal comes closest to tracking the actual drift using only unlabeled data, with a high robustness to false alarms. The margin density approach is robust not only to changes in the irrelevant features, but also to changes in the relevant features, which are not critical to a robust classifier's performance. 

The \textit{Uncertain} and the \textit{HD} metrics, presented here, are representations of two paradigms of unlabeled drift detection techniques in literature. The \textit{HD} metric represents the fundamental behavior of the feature distribution tracking techniques (Section~\ref{sec:lr_multivariate}) \citep{lee2012detection,ditzler2011hellinger, kuncheva2014pca,qahtan2015pca}, while the \textit{Uncertainty} metric represents behavior of the model based posterior probabilities tracking techniques (Section~\ref{sec:lr_model}) \citep{dries2009adaptive,lindstrom2013drift,dredze2010we,zliobaite2010change}
. Although the actual usage and the implementation of the metrics are nuanced and different in literature, our purpose here is to demonstrate the underlying principle that makes the margin density signal more robust to false alarms. The purpose of this section is to provide motivation for future work on improving reliability of unlabeled drift detection techniques. The generation and analysis of real world datasets, with similar characteristics as the synthetic data presented here, could also be a useful contribution to better enable further research in this area.

\section{Conclusion and future work}
\label{sec:conclusion}

In this paper, the Margin Density Drift Detection (MD3) methodology is presented, which is capable of reliably detecting concept drift from unlabeled streaming data. 
The proposed methodology uses the number of samples in a classifier's region of uncertainty (margin), as a metric for detecting drift. A significant deviation in the margin density metric was used to signal the need for labeled samples, for subsequent verification and retraining of the classifier. MD3 was shown to be independent of the classification algorithm used, robust to stray changes in data distribution and a reliable substitute to supervised drift detectors. Furthermore, the development of MD3 as an incremental drift detection algorithm, makes it amenable for usage in a variety of data stream environments. 

Experimental analysis was performed on 6 drift induced datasets, 4 real world concept drift datasets from the cybersecurity domain, and on 2 benchmark concept drift datasets. The results indicated high detection rate and prediction performance, coupled with low false alarm rate for the MD3 approach. The margin density approach, when tested on 6 drift induced datasets, resulted in a $<$1\% difference in the accuracy, on average,  compared to a fully labeled drift detection methodology. When tested on the 6 real world cybersecurity datasets, the average difference was 1.05\%, indicating its efficacy to be used as a surrogate for fully labeled approaches. When compared with a state of the art unlabeled feature tracking approach - The Hellinger Distance Drift Detection Methodology (HDDDM), the MD3 algorithm resulted in fewer false alarms and a smaller labeling percentage (8.7\% lesser on average), for the same ($\Delta$=+0.66\%) accuracy.   Additionally, experimentation using classifiers with explicit margins and those without, demonstrated no significant differences, indicating the generality of the margin density approach as a model independent signal of change. 

Evaluation of the MD3 approach on cybersecurity datasets from the domains of - spam detection, phishing websites and network intrusion detection, demonstrates the applicability of the proposed work to detection of adversarial activity. While the MD3 approach is general and domain independent in its applicability, we particularly highlight benefits in adversarial domains, which we envision can immediately benefit from reduced number of false alarms and increased reliability of change detection. The approach used 88.8\% less labeling than supervised drift detection approaches, while producing statistically equivalent classification performance. Also, the MD3 approach outperformed the unsupervised HDDDM approach, by reducing the false alarm rate by 71\%, on average. Research in adversarial drift handling systems \citep{kantchelian2013approaches} can utilize this approach to develop unsupervised dynamic machines, capable of constantly changing, to subvert adversarial activity. 

The MD3 approach saves labeling expenditure by signaling change only when it can lead to a drop in the predictive performance of the classifier. Future work will concentrate on label efficient techniques for relearning after a drift has been detected.  Using the margin density information along with active learning strategies \citep{zliobaite2014active}, to selectively label samples for the confirmation and retraining phase, warrants further research. This would allow for additional savings in terms of reduced labeling, in the online classification process.




\section*{References}

\bibliographystyle{elsarticle-harv} 
\bibliography{references}

\begin{thebibliography}{62}
\expandafter\ifx\csname natexlab\endcsname\relax\def\natexlab#1{#1}\fi
\expandafter\ifx\csname url\endcsname\relax
  \def\url#1{\texttt{#1}}\fi
\expandafter\ifx\csname urlprefix\endcsname\relax\def\urlprefix{URL }\fi

\bibitem[{Ahn et~al.(2007)Ahn, Moon, Fazzari, Lim, Chen, and
  Kodell}]{ahn2007classification}
Ahn, H., Moon, H., Fazzari, M.~J., Lim, N., Chen, J.~J., Kodell, R.~L., 2007.
  Classification by ensembles from random partitions of high-dimensional data.
  Computational Statistics \& Data Analysis 51~(12), 6166--6179.

\bibitem[{Bach and Maloof(2008)}]{bach2008paired}
Bach, S.~H., Maloof, M.~A., 2008. Paired learners for concept drift. In: Eighth
  IEEE International Conference on Data Mining (ICDM). IEEE, pp. 23--32.

\bibitem[{Baena-Garc{\i}a et~al.(2006)Baena-Garc{\i}a, del Campo-{\'A}vila,
  Fidalgo, Bifet, Gavalda, and Morales-Bueno}]{baena2006early}
Baena-Garc{\i}a, M., del Campo-{\'A}vila, J., Fidalgo, R., Bifet, A., Gavalda,
  R., Morales-Bueno, R., 2006. Early drift detection method. In: Fourth
  international workshop on knowledge discovery from data streams. Vol.~6. pp.
  77--86.

\bibitem[{Barreno et~al.(2010)Barreno, Nelson, Joseph, and
  Tygar}]{barreno2010security}
Barreno, M., Nelson, B., Joseph, A.~D., Tygar, J., 2010. The security of
  machine learning. Machine Learning 81~(2), 121--148.

\bibitem[{Bifet and Gavalda(2007)}]{bifet2007learning}
Bifet, A., Gavalda, R., 2007. Learning from time-changing data with adaptive
  windowing. In: SDM. Vol.~7. SIAM.

\bibitem[{Bifet et~al.(2010)Bifet, Holmes, Pfahringer, and
  Frank}]{bifet2010fast}
Bifet, A., Holmes, G., Pfahringer, B., Frank, E., 2010. Fast perceptron
  decision tree learning from evolving data streams. In: Pacific-Asia
  Conference on Knowledge Discovery and Data Mining. Springer, pp. 299--310.

\bibitem[{Bryll et~al.(2003)Bryll, Gutierrez-Osuna, and Quek}]{bry}
Bryll, R., Gutierrez-Osuna, R., Quek, F., 2003. Attribute bagging: improving
  accuracy of classifier ensembles by using random feature subsets. Pattern
  recognition 36~(6), 1291--1302.

\bibitem[{Burnap and Williams(2016)}]{burnap2016us}
Burnap, P., Williams, M.~L., 2016. Us and them: identifying cyber hate on
  twitter across multiple protected characteristics. EPJ Data Science 5~(1), 1.

\bibitem[{Cha(2007)}]{cha2007comprehensive}
Cha, S.-H., 2007. Comprehensive survey on distance/similarity measures between
  probability density functions. City 1~(2), 1.

\bibitem[{Chang and Lin(2011)}]{chang2011libsvm}
Chang, C.-C., Lin, C.-J., 2011. Libsvm: a library for support vector machines.
  ACM Transactions on Intelligent Systems and Technology (TIST) 2~(3), 27.

\bibitem[{Chinavle et~al.(2009)Chinavle, Kolari, Oates, and
  Finin}]{chinavle2009ensembles}
Chinavle, D., Kolari, P., Oates, T., Finin, T., 2009. Ensembles in adversarial
  classification for spam. In: Proceedings of the 18th ACM conference on
  Information and knowledge management. ACM, pp. 2015--2018.

\bibitem[{Cover and Hart(1967)}]{cover1967nearest}
Cover, T.~M., Hart, P.~E., 1967. Nearest neighbor pattern classification. IEEE
  Transactions on Information Theory 13~(1), 21--27.

\bibitem[{da~Costa et~al.(2016)da~Costa, Rios, and de~Mello}]{da2016using}
da~Costa, F., Rios, R., de~Mello, R., 2016. Using dynamical systems tools to
  detect concept drift in data streams. Expert Systems with Applications 60,
  39--50.

\bibitem[{Dem{\v{s}}ar(2006)}]{demvsar2006statistical}
Dem{\v{s}}ar, J., 2006. Statistical comparisons of classifiers over multiple
  data sets. Journal of Machine learning research 7~(Jan), 1--30.

\bibitem[{Dietterich(2000)}]{dietterich2000ensemble}
Dietterich, T.~G., 2000. Ensemble methods in machine learning. In: Multiple
  classifier systems. Springer, pp. 1--15.

\bibitem[{Ditzler and Polikar(2011)}]{ditzler2011hellinger}
Ditzler, G., Polikar, R., 2011. Hellinger distance based drift detection for
  nonstationary environments. In: IEEE Symposium on Computational Intelligence
  in Dynamic and Uncertain Environments (CIDUE). IEEE, pp. 41--48.

\bibitem[{Dredze et~al.(2010)Dredze, Oates, and Piatko}]{dredze2010we}
Dredze, M., Oates, T., Piatko, C., 2010. We're not in kansas anymore: detecting
  domain changes in streams. In: Proceedings of the 2010 Conference on
  Empirical Methods in Natural Language Processing. Association for
  Computational Linguistics, pp. 585--595.

\bibitem[{Dries and R{\"u}ckert(2009)}]{dries2009adaptive}
Dries, A., R{\"u}ckert, U., 2009. Adaptive concept drift detection. Statistical
  Analysis and Data Mining 2~(5-6), 311--327.

\bibitem[{Duch et~al.(2004)Duch, Wieczorek, Biesiada, and
  Blachnik}]{duch2004comparison}
Duch, W., Wieczorek, T., Biesiada, J., Blachnik, M., 2004. Comparison of
  feature ranking methods based on information entropy. In: IEEE International
  Joint Conference on Neural Networks. Vol.~2. IEEE, pp. 1415--1419.

\bibitem[{Faria et~al.(2013)Faria, Gama, and Carvalho}]{faria2013novelty}
Faria, E.~R., Gama, J., Carvalho, A.~C., 2013. Novelty detection algorithm for
  data streams multi-class problems. In: Proceedings of the 28th Annual ACM
  Symposium on Applied Computing. ACM, pp. 795--800.

\bibitem[{Farid et~al.(2013)Farid, Zhang, Hossain, Rahman, Strachan, Sexton,
  and Dahal}]{farid2013adaptive}
Farid, D.~M., Zhang, L., Hossain, A., Rahman, C.~M., Strachan, R., Sexton, G.,
  Dahal, K., 2013. An adaptive ensemble classifier for mining concept drifting
  data streams. Expert Systems with Applications 40~(15), 5895--5906.

\bibitem[{Gama et~al.(2004)Gama, Medas, Castillo, and
  Rodrigues}]{gama2004learning}
Gama, J., Medas, P., Castillo, G., Rodrigues, P., 2004. Learning with drift
  detection. In: Advances in artificial intelligence--SBIA 2004. Springer, pp.
  286--295.

\bibitem[{Gao et~al.(2007)Gao, Fan, and Han}]{gao2007appropriate}
Gao, J., Fan, W., Han, J., 2007. On appropriate assumptions to mine data
  streams: Analysis and practice. In: Seventh IEEE International Conference on
  Data Mining (ICDM). IEEE, pp. 143--152.

\bibitem[{Goncalves et~al.(2014)Goncalves, de~Carvalho~Santos, Barros, and
  Vieira}]{goncalves2014comparative}
Goncalves, P.~M., de~Carvalho~Santos, S.~G., Barros, R.~S., Vieira, D.~C.,
  2014. A comparative study on concept drift detectors. Expert Systems with
  Applications 41~(18), 8144--8156.

\bibitem[{Harel et~al.(2014)Harel, Mannor, El-Yaniv, and
  Crammer}]{harel2014concept}
Harel, M., Mannor, S., El-Yaniv, R., Crammer, K., 2014. Concept drift detection
  through resampling. In: Proceedings of the 31st International Conference on
  Machine Learning (ICML-14). pp. 1009--1017.

\bibitem[{Haussler(1990)}]{haussler1990probably}
Haussler, D., 1990. Probably approximately correct learning. University of
  California, Santa Cruz, Computer Research Laboratory.

\bibitem[{Hayat and Hashemi(2010)}]{hayat2010dct}
Hayat, M.~Z., Hashemi, M.~R., 2010. A dct based approach for detecting novelty
  and concept drift in data streams. In: International Conference of Soft
  Computing and Pattern Recognition (SoCPaR). IEEE, pp. 373--378.

\bibitem[{Kantchelian et~al.(2013)Kantchelian, Afroz, Huang, Islam, Miller,
  Tschantz, Greenstadt, Joseph, and Tygar}]{kantchelian2013approaches}
Kantchelian, A., Afroz, S., Huang, L., Islam, A.~C., Miller, B., Tschantz,
  M.~C., Greenstadt, R., Joseph, A.~D., Tygar, J., 2013. Approaches to
  adversarial drift. In: Proceedings of the ACM workshop on Artificial
  intelligence and security. ACM, pp. 99--110.

\bibitem[{Katakis et~al.(2009)Katakis, Tsoumakas, Banos, Bassiliades, and
  Vlahavas}]{katakis2009adaptive}
Katakis, I., Tsoumakas, G., Banos, E., Bassiliades, N., Vlahavas, I., 2009. An
  adaptive personalized news dissemination system. Journal of Intelligent
  Information Systems 32~(2), 191--212.

\bibitem[{Katakis et~al.(2010)Katakis, Tsoumakas, and
  Vlahavas}]{katakis2010tracking}
Katakis, I., Tsoumakas, G., Vlahavas, I., 2010. Tracking recurring contexts
  using ensemble classifiers: an application to email filtering. Knowledge and
  Information Systems 22~(3), 371--391.

\bibitem[{Kohavi et~al.(1995)}]{kohavi1995study}
Kohavi, R., et~al., 1995. A study of cross-validation and bootstrap for
  accuracy estimation and model selection. In: Ijcai. Vol.~14. pp. 1137--1145.

\bibitem[{Krempl et~al.(2014)Krempl, {\v{Z}}liobaite, Brzezi{\'n}ski,
  H{\"u}llermeier, Last, Lemaire, Noack, Shaker, Sievi, Spiliopoulou,
  et~al.}]{krempl2014open}
Krempl, G., {\v{Z}}liobaite, I., Brzezi{\'n}ski, D., H{\"u}llermeier, E., Last,
  M., Lemaire, V., Noack, T., Shaker, A., Sievi, S., Spiliopoulou, M., et~al.,
  2014. Open challenges for data stream mining research. ACM SIGKDD
  explorations newsletter 16~(1), 1--10.

\bibitem[{Kuncheva(2008)}]{kuncheva2008classifier}
Kuncheva, L.~I., 2008. Classifier ensembles for detecting concept change in
  streaming data: Overview and perspectives. In: 2nd Workshop SUEMA. Vol. 2008.
  pp. 5--10.

\bibitem[{Kuncheva and Faithfull(2014)}]{kuncheva2014pca}
Kuncheva, L.~I., Faithfull, W.~J., 2014. Pca feature extraction for change
  detection in multidimensional unlabeled data. IEEE Transactions on Neural
  Networks and Learning Systems 25~(1), 69--80.

\bibitem[{Lee and Magoules(2012)}]{lee2012detection}
Lee, J., Magoules, F., 2012. Detection of concept drift for learning from
  stream data. In: IEEE 14th International Conference on High Performance
  Computing and Communication \& 2012 IEEE 9th International Conference on
  Embedded Software and Systems (HPCC-ICESS). IEEE, pp. 241--245.

\bibitem[{Lichman(2013)}]{Lichman:2013}
Lichman, M., 2013. {UCI} machine learning repository.
\newline\urlprefix\url{http://archive.ics.uci.edu/ml}

\bibitem[{Lindstrom et~al.(2010)Lindstrom, Delany, and
  Mac~Namee}]{lindstrom2010handling}
Lindstrom, P., Delany, S.~J., Mac~Namee, B., 2010. Handling concept drift in a
  text data stream constrained by high labelling cost. In: Twenty-Third
  International FLAIRS Conference.

\bibitem[{Lindstrom et~al.(2013)Lindstrom, Mac~Namee, and
  Delany}]{lindstrom2013drift}
Lindstrom, P., Mac~Namee, B., Delany, S.~J., 2013. Drift detection using
  uncertainty distribution divergence. Evolving Systems 4~(1), 13--25.

\bibitem[{Lughofer et~al.(2016)Lughofer, Weigl, Heidl, Eitzinger, and
  Radauer}]{lughofer2016recognizing}
Lughofer, E., Weigl, E., Heidl, W., Eitzinger, C., Radauer, T., 2016.
  Recognizing input space and target concept drifts in data streams with
  scarcely labeled and unlabelled instances. Information Sciences 355,
  127--151.

\bibitem[{Masud et~al.(2011)Masud, Gao, Khan, Han, and
  Thuraisingham}]{masud2011classification}
Masud, M.~M., Gao, J., Khan, L., Han, J., Thuraisingham, B., 2011.
  Classification and novel class detection in concept-drifting data streams
  under time constraints. IEEE TKDE 23~(6), 859--874.

\bibitem[{Nishida and Yamauchi(2007)}]{nishida2007detecting}
Nishida, K., Yamauchi, K., 2007. Detecting concept drift using statistical
  testing. In: Discovery Science. Springer, pp. 264--269.

\bibitem[{Page(1954)}]{page1954continuous}
Page, E., 1954. Continuous inspection schemes. Biometrika 41~(1/2), 100--115.

\bibitem[{Pedregosa et~al.(2011)Pedregosa, Varoquaux, Gramfort, Michel,
  Thirion, Grisel, Blondel, Prettenhofer, Weiss, Dubourg, Vanderplas, Passos,
  Cournapeau, Brucher, Perrot, and Duchesnay}]{scikit-learn}
Pedregosa, F., Varoquaux, G., Gramfort, A., Michel, V., Thirion, B., Grisel,
  O., Blondel, M., Prettenhofer, P., Weiss, R., Dubourg, V., Vanderplas, J.,
  Passos, A., Cournapeau, D., Brucher, M., Perrot, M., Duchesnay, E., 2011.
  Scikit-learn: Machine learning in {P}ython. Journal of Machine Learning
  Research 12, 2825--2830.

\bibitem[{Qahtan et~al.(2015)Qahtan, Alharbi, Wang, and Zhang}]{qahtan2015pca}
Qahtan, A.~A., Alharbi, B., Wang, S., Zhang, X., 2015. A pca-based change
  detection framework for multidimensional data streams: Change detection in
  multidimensional data streams. In: Proc. of the 21th ACM SIGKDD ICKDDM. ACM,
  pp. 935--944.

\bibitem[{Quinlan(1993)}]{quinlan1993c4}
Quinlan, J.~R., 1993. C4. 5: Programming for machine learning. Morgan
  Kauffmann.

\bibitem[{Ross et~al.(2012)Ross, Adams, Tasoulis, and
  Hand}]{ross2012exponentially}
Ross, G.~J., Adams, N.~M., Tasoulis, D.~K., Hand, D.~J., 2012. Exponentially
  weighted moving average charts for detecting concept drift. Pattern
  Recognition Letters 33~(2), 191--198.

\bibitem[{Ryu et~al.(2012)Ryu, Kantardzic, Kim, and Khil}]{ryu2012efficient}
Ryu, J.~W., Kantardzic, M.~M., Kim, M.-W., Khil, A.~R., 2012. An efficient
  method of building an ensemble of classifiers in streaming data. In: Big data
  analytics. Springer, pp. 122--133.

\bibitem[{Schmidt et~al.(1995)Schmidt, Siegel, and
  Srinivasan}]{schmidt1995chernoff}
Schmidt, J.~P., Siegel, A., Srinivasan, A., 1995. Chernoff-hoeffding bounds for
  applications with limited independence. SIAM Journal on Discrete Mathematics
  8~(2), 223--250.

\bibitem[{Sethi and Kantardzic(2015)}]{sethi2015don}
Sethi, T.~S., Kantardzic, M., 2015. Don't pay for validation: Detecting drifts
  from unlabeled data using margin density. Procedia Computer Science 53,
  103--112.

\bibitem[{Sethi et~al.(2016{\natexlab{a}})Sethi, Kantardzic, and
  Arabmakki}]{tsethi}
Sethi, T.~S., Kantardzic, M., Arabmakki, E., 2016{\natexlab{a}}. Monitoring
  classification blindspots to detect drifts from unlabeled data. In: 17th IEEE
  International Conference on Information Reuse and Integration (IRI). IEEE.

\bibitem[{Sethi et~al.(2016{\natexlab{b}})Sethi, Kantardzic, and
  Hu}]{sethi2016grid}
Sethi, T.~S., Kantardzic, M., Hu, H., 2016{\natexlab{b}}. A grid density based
  framework for classifying streaming data in the presence of concept drift.
  Journal of Intelligent Information Systems 46~(1), 179--211.

\bibitem[{Settles(2012)}]{settles2010active}
Settles, B., 2012. Active learning. Synthesis Lectures on Artificial
  Intelligence and Machine Learning 6~(1), 1--114.

\bibitem[{Skurichina and Duin(2002)}]{skurichina2002bagging}
Skurichina, M., Duin, R.~P., 2002. Bagging, boosting and the random subspace
  method for linear classifiers. Pattern Analysis \& Applications 5~(2),
  121--135.

\bibitem[{Smutz and Stavrou(2016)}]{smutz2016tree}
Smutz, C., Stavrou, A., 2016. When a tree falls: Using diversity in ensemble
  classifiers to identify evasion in malware detectors. In: NDSS Symposium.

\bibitem[{Sobhani and Beigy(2011)}]{sobhani2011new}
Sobhani, P., Beigy, H., 2011. New drift detection method for data streams.
  Springer.

\bibitem[{Spinosa et~al.(2007)Spinosa, de~Leon F~de Carvalho, and
  Gama}]{spinosa2007olindda}
Spinosa, E.~J., de~Leon F~de Carvalho, A.~P., Gama, J., 2007. Olindda: A
  cluster-based approach for detecting novelty and concept drift in data
  streams. In: Proceedings of the 2007 ACM symposium on Applied computing. ACM,
  pp. 448--452.

\bibitem[{Tavallaee et~al.(2009)Tavallaee, Bagheri, Lu, and
  Ghorbani}]{tavallaee2009detailed}
Tavallaee, M., Bagheri, E., Lu, W., Ghorbani, A.-A., 2009. A detailed analysis
  of the kdd cup 99 data set. In: Proceedings of the Second IEEE Symposium on
  Computational Intelligence for Security and Defence Applications 2009.

\bibitem[{Wang(2015)}]{wang2015robust}
Wang, F., 2015. Robust and adversarial data mining. Sydney Digital Theses (Open
  Access).

\bibitem[{Wang and Abraham(2015)}]{wang2015concept}
Wang, H., Abraham, Z., 2015. Concept drift detection for streaming data. In:
  International Joint Conference on Neural Networks (IJCNN). IEEE, pp. 1--9.

\bibitem[{Wu et~al.(2014)Wu, Zhu, Wu, and Ding}]{wu2014data}
Wu, X., Zhu, X., Wu, G.-Q., Ding, W., 2014. Data mining with big data.
  Knowledge and Data Engineering, IEEE Transactions on 26~(1), 97--107.

\bibitem[{Zliobaite(2010)}]{zliobaite2010change}
Zliobaite, I., 2010. Change with delayed labeling: when is it detectable? In:
  IEEE International Conference on Data Mining Workshops (ICDMW). IEEE, pp.
  843--850.

\bibitem[{Zliobaite et~al.(2014)Zliobaite, Bifet, Pfahringer, and
  Holmes}]{zliobaite2014active}
Zliobaite, I., Bifet, A., Pfahringer, B., Holmes, G., 2014. Active learning
  with drifting streaming data. IEEE Transactions on Neural Networks and
  Learning Systems 25~(1), 27--39.

\end{thebibliography}




\end{document}